\documentclass[10pt,a4paper,two column]{article}
 \pdfoutput=1
\usepackage{amsmath,amsfonts,amssymb}
\usepackage{url}
\usepackage[utf8]{inputenc}
\usepackage[font=footnotesize]{caption}
\usepackage{microtype} 
\usepackage{booktabs} 
\usepackage{pseudocode}
\usepackage{wrapfig}
\usepackage{float}
\usepackage{graphicx}
\newfloat{algorithm}{t}{lop}
\usepackage{mathptmx}

\newcommand{\x}{\mathbf{ x}}
\newcommand{\w}{\mathbf{ w}}
\newcommand{\R}{\mathbb{R}}
\newcommand{\RelSim}{\text{RelSim}}
\newcommand{\relsim}{\mbox{RelSim}}

\newcommand{\tet}{\pmb{\theta}}
\newcommand{\ind}{\boldsymbol{I}}
\newcommand{\opt}{\mathrm{opt}}
\newcommand{\argmin}[1]{\ensuremath{\operatorname{arg\,}\operatornamewithlimits{min}_{#1}}}
\newcommand{\argmax}[1]{\ensuremath{\operatorname{arg\,}\operatornamewithlimits{max}_{#1}}}

\title{Optimum Reject Options for Prototype-based Classification}

\author{ \small
Lydia~Fischer$^{1,2}$,  Barbara~Hammer$^{1}$,  and~Heiko~Wersing$^{2}$
\thanks{L. Fischer acknowledges funding by the CoR-Lab Research Institute for Cognition and Robotics and gratefully acknowledges the financial support from the Honda Research Institute Europe. B. Hammer gratefully acknowledges funding by the CITEC center of excellence.}\\
\small
 1 -- Bielefeld University, Universit\"atsstra\ss e 25, 336615 Bielefeld, Germany\\
 \small
 2 -- HONDA Research Institute Europe, Carl-Legien-Stra\ss e 30, 63065 Offenbach, Germany
}

\begin{document}

\maketitle
\small
\textit{Abstract} -- \textbf{We analyse optimum reject strategies for prototype-based classifiers and real-valued rejection measures, using the distance of a data point to the closest prototype or probabilistic counterparts.
We compare reject schemes with global thresholds, and local thresholds for the Voronoi cells of the classifier.
For the latter, we develop a polynomial-time algorithm to compute optimum thresholds based on a dynamic programming scheme, and we propose an intuitive linear time, memory efficient approximation thereof with competitive accuracy.
Evaluating the performance in various benchmarks, we conclude that local reject options are beneficial in particular for  simple prototype-based classifiers, while the improvement is less pronounced for advanced models.
For the latter, an accuracy-reject curve which is comparable to support vector machine classifiers with state of the art reject options can be reached.}

\vspace*{0.05cm}
\normalsize
\section{Introduction}
\subsection{Motivation}
Classification constitutes one of the standard application scenarios for machine learning techniques: 
Its application ranges from automated digit recognition up to fraud detection, and numerous machine learning models are readily available for this task \cite{Bishop2006}.
Often, besides the overall classification accuracy, the flexibility of the classification model to handle uncertain predictions plays an important role.
Techniques which provide a level of certainty together with the predicted class label can trade classification security for a partial prediction of the labels; in the latter case, data for which the prediction is insecure are rejected.
In particular applications which require a life long learning or an adaptation to changing conditions  benefit from such flexible classification models \cite{heiko}.
Moreover, in safety critical areas such as driver assistance systems, health care, or biomedical data analysis, the information about the certainty of the classification is almost as important as the class label itself.
Further tests or expert opinions can be consulted for uncertain classification to avoid critical effects of a misclassification for instance in health care.
For driver assistance systems, a high degree of uncertainty can result in turning off the assistance system and passing the responsibility back to the human driver.
In all settings, the possibility of a machine learning classifier to reject a classification in case of a low classification confidence is crucial.

Reject options have been pioneered by the formal framework as investigated in the approach \cite{Chow1970}: 
If the costs for a misclassification versus a reject are known, one can design an optimum reject threshold based on the probability of misclassification.
In practice, however, the exact probability  of a misclassification is generally unknown.
Hence further research addresses the question whether reject options can be based on plugin rules where only empirical estimates of the misclassification probability are used \cite{herbei:2005}.
Still, these  formalisations rely on consistent probability estimates, which are often not present for given classifiers.
Further, rejection and misclassification costs need to be known and constant, which is not necessarily the case in particular for online settings. 
Thus, these settings deal with an idealised modelling and are not necessarily applicable for efficient, possibly deterministic classifiers in complex scenarios.

Some machine learning classifiers allow an intuitive incorporation of reject options.
Naturally, probabilistic classifiers can directly be plugged into the framework as analysed in \cite{herbei:2005}.
In particular, probabilistic classifiers provide confidence values for which the given scaling is meaningful, provided the probability estimation is correct.
The latter is often not the case since the model assumptions need not hold, the model often relies on simplified assumptions, or model priors are chosen based on computational feasibility rather than the (unknown) underlying truth.
Further, the inference of exact probabilistic models is not always feasible, depending on the type and size of data and the available ground truth.

One principled alternative to enhance given classifiers by confidence values is offered by bootstrapping \cite{Bishop2006}.
This, however, requires repeated training with the available training data, such that it displays a high computational and memory complexity.
Further, it is not applicable for online settings where data are not necessarily independent and identically distributed.
For online learning, the theory of conformal prediction has caused quite some interest recently \cite{cp1,cp2}. 
The formalism is based on a so called non-conformity measure; having chosen a suitable criterion, it provides a statistically well founded theory to estimate the confidence of a classification in online settings, where data have to fulfil the weaker property of interchangeability only rather than being i.\,i.\,d., see \cite{cp1,cp2}.
In practice, however, the choice of the non-conformity measure is very critical and suboptimal choices do not lead to meaningful results.
Further, the original approach is very time consuming since it requires the re-training of the model in a leave-one-out fashion.
Albeit efficient approximations exist for some classifiers such as prototype-based models, the formal guarantees usually do no longer hold for the latter \cite{PRL14Zhuetal}.

There have been attempts to accompany powerful deterministic classifiers by efficient ways of confidence estimation.
One popular example is given for the support vector machine (SVM), see the approach \cite{Platt99} for two-class classification and the work \cite{Wu2004} for extensions towards multiple classes.
These techniques are implemented e.\,g.\ in the popular LIBSVM \cite{libsvm}.
In this article, we are interested in an alternative classification paradigm:  
Prototype-based models which represent classes in terms of typical representatives and thus allow a direct inspection of the classifier.
This feature has contributed to an increasing popularity of these models in particular in the biomedical domain, see e.\,g.\ \cite{Biehl03072014,Alegre2013525,Huber201291,arltbiehl2011jcem}, by offering an elegant  representation which lends itself to model interpretability in a natural way \cite{interesann,Freitas:2014:CCM:2594473.2594475,matters}.
Further, the representation of models in terms of few representative prototypes has proved useful when dealing with online scenarios or big data sets \cite{esann14Hametal,Kietzmann08incrementalgrlvq,heiko}.
While some approaches exist to accompany nearest neighbour based classification or Gaussian mixture models (GMM) by confidence estimations
\cite{ICPR00VOLII48,Huelal09}, first reject options for discriminative prototype-based methods such as learning vector quantisation have only recently been proposed \cite{esann,wsom}.
In this article, we will built on the insights as gained in the recent approaches \cite{esann,wsom}, and we will investigate how to optimally set the thresholds within intuitive reject schemes for prototype-based techniques.

While the threshold selection strategies which we will investigate can be used for any prototype-based classification scheme, we will focus on the popular supervised classification technique learning vector quantisation (LVQ) and its recent more fundamental mathematical derivatives 
\cite{Kohonen1989,SeoObermayer2003,SchneiderBH09a,schneider2010regularization}.
LVQ constitutes a powerful and efficient meth\-od for multi-class classification tasks which, due to its simple representation of models in terms of prototypes, is particularly suited for interpretability,  online scenarios or life long learning \cite{heiko}.
While classical LVQ models mostly rely on  heuristics, modern variants are based on cost-functions such as  generalized LVQ (GLVQ) \cite{SatoYamada1995}, or the full probabilistic model robust soft LVQ (RSLVQ) \cite{SeoObermayer2003}.
LVQ classifiers can be accompanied by strong guarantees concerning their generalization performance and learning dynamics \cite{SchneiderBH09a,2007dynamics}.
One particular success story links LVQ classifiers to metric learners: 
These  enrich the classifier by feature weighting terms  which opens the way towards a more flexible classification scheme, increased model interpretability, and even a simultaneous visualisation
of the classifier \cite{SchneiderBH09a,schneider2010regularization,biehl}.
Further, recent LVQ variants address the setting of complex, possibly non-euclidean data which are described by pairwise similarities or dissimilarities only \cite{Neucom13Hametal}.
Apart from the probabilistic model RSLVQ, these classifiers are often deterministic and do not provide a confidence of the classification. Further, also for RSLVQ, the correctness of the probability estimate is not clear since the model is not designed in order to correctly model the data probability but the conditional label probability only \cite{2007dynamics,wsom}.

In this contribution, building on the results as recently published in \cite{esann} which proposes different real-valued certainty measures suitable for an integration in a reject option, we investigate how to devise optimum reject strategies for LVQ type classifiers, putting a particular emphasis on the choice of the threshold for a reject.
In particular, we are interested in efficient, online-computable reject options for LVQ classifiers  and their behaviour in comparison to mathematically well founded statistical models and the SVM \cite{svm1, svm2}.
We will compare reject strategies based on one global reject threshold, and local reject thresholds which take into account the Voronoi tessellation of the space induced by the prototypes. 
For the latter, we present an optimum computation scheme how to set the threshold for the different Voronoi cells, and we also propose a time and memory efficient approximation thereof which lends itself to online scenarios. We evaluate the techniques extensively using different benchmark data and different types of LVQ classifiers, also providing a comparison to classification with rejection based on SVM. Further, we demonstrate the suitability of the devised technique for  a real life example from a medical domain.

This contribution is structured as follows:
In section \ref{literatur} we give an overview about existing methods to enhance classifiers by reject options.
Afterwards in section \ref{algos_data} we explain the LVQ training algorithms that we use in our experiments and introduce in section \ref{reject_option} the basic schemes how to reject based on global or local thresholds.
Thereby, we develop a polynomial time scheme based on dynamic programming (DP), that allows an optimum choice of local thresholds, as well as a time and memory efficient greedy approximation thereof. 
Further, we present in section \ref{measures} suitable certainty measures that can be plugged into these reject schemes. 
In the experiments section \ref{experiments}, we test the techniques using different benchmarks and LVQ learning schemes. We illustrate the suitability of the methods, whereby we put a particular emphasis on the comparison of local versus global reject schemes, the comparison of an optimum computation of local thresholds by means of DP versus an efficient greedy scheme, and a comparison of the proposed reject schemes for highly flexible LVQ classifiers with state of the art reject options which accompany an SVM.
For all experiments, evaluation will rely on the full accuracy-reject curve as proposed in the approach \cite{NadeemZH10}.

\subsection{Related Work}\label{literatur}
The following section summarises the state of the art for reject options and accompanying certainty measures in supervised learning. 
The approach  \cite{ICPR00VOLII48} highlights two main reasons for rejection:
\begin{itemize}
	\item Ambiguity:  It is not clear how to classify the data point, e.\,g.\ the point is close to at least one decision boundary, or it lies in a region where at least two classes are overlapping.
	\item Outliers: The data point is dissimilar to any already seen data
	point, e.\,g.\  it is caused by noise or it is an instance of a yet unseen class or cluster.
\end{itemize}
There exist several approaches which explicitly address one of these reasons or a combination of both.
Mostly, reject options are based on a measure which provides a certainty value about whether a given data point is correctly classified.
In the following,  we distinguish measures which are based on heuristics and approaches which are based on estimates of misclassification probabilities or confidences.
We primarily focus on techniques which have been proposed for distance-based classifiers and similar due to their similarity to LVQ techniques.

%
\paragraph{Heuristic Measures:}
\begin{wrapfigure}[11]{r}{0.45\columnwidth}
\centering
\vspace*{-0.4cm}
\includegraphics[width=0.4\columnwidth]{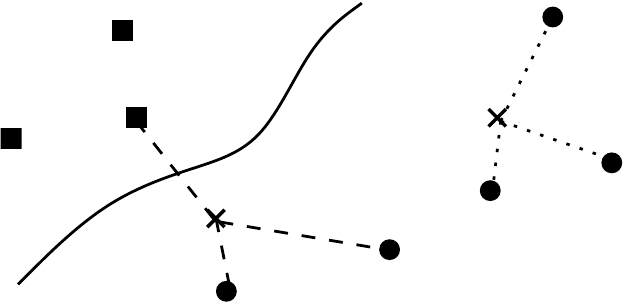} 
\caption{Sketch of a possible $k$-NN reject scheme ($k=3$). Different symbols indicate different classes. Classification of the left data point ($\times$) is more uncertain than the right one ($\times$) because all neighbours are of the same class for the latter. }
\label{fig:knn_reject}
\end{wrapfigure}
For $k$-nearest neighbour ($k$-NN) \cite{knn} approaches a variety of  simple certainty measures exist using a neighbourhood of a given data point \cite{Delanyelal05, Huelal09}.
These measures rely on the correlation of the label of the data point and its neighbours (cp. Fig.~\ref{fig:knn_reject}).
In these approaches, several different realisations and combinations of the counting have been compared, leading to the result that an ensemble measure largely rises the stability of the single measures.
The approach \cite{SugiyamaB13} focusses on effective outlier detection, relying on the  distances of a new data point from elements of a randomly chosen subset of the given data.
An outlier score is then given by the smallest distance.
The resulting method outperforms state of the art approaches such as proposed in \cite{RamaswamyRS00} in efficiency and accuracy.

Sousa \& Cardoso \cite{SousaC13} introduce a reject option which identifies ambiguous regions in binary classifications.
Their approach is based on a data replication method. 
An advantage of the proposed strategy is given by the fact that no reject threshold has to be set externally, rather the technique itself  provides a suitable cutoff.

The approach \cite{StefanoSV00} addresses different neural network architectures including multi-layer perceptrons, learning vector quantisation, and probabilistic neural networks.
Here an effectiveness function is introduced taking different costs for rejection and classification errors into account, very similar to the loss function as considered in \cite{Chow1970,herbei:2005}.
Then, different rejection measures based on the activation of the output neurons are investigated. 

\begin{wrapfigure}[13]{r}{0.5\columnwidth}
\centering
\includegraphics[width=0.39\columnwidth]{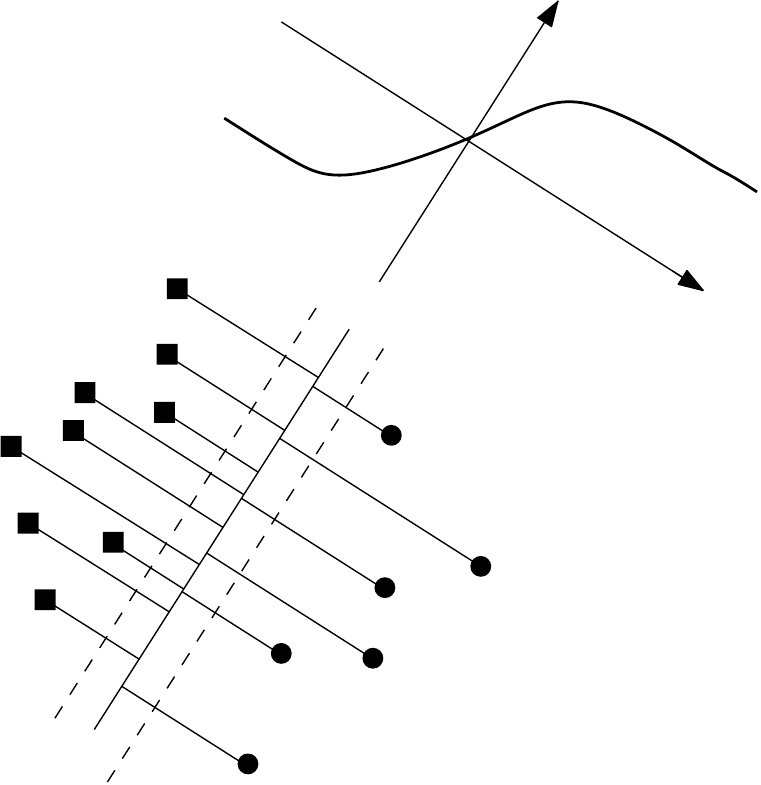} 
\caption{Sketch of a binary classification setting in SVM. A Sigmoid is fitted against the values of the bins of the distances from the data points to the separating hyperplane.}
\label{fig:svm_reject}
\end{wrapfigure}
A very popular approach to turn the activity provided by a binary SVM into an approximation of a classification confidence measure has been proposed by Platt \cite{Platt99}.
The certainty measure is based on the distance of a data point to the decision border, i.\,e.\ the activation of an SVM classifier. 
By means of a non-linear sigmoidal function, the distance is transformed to a confidence value. 
Thereby, the parameters of the sigmoidal are fitted on the given training data (Fig.~\ref{fig:svm_reject}).
A transfer of this method for multi-class tasks is provided by Wu et al. \cite{Wu2004} and it is implemented in the popular LIBSVM toolbox \cite{libsvm}.

%
\paragraph{Probabilistic Measures:}
There exist several approaches which more closely rely on an explicit probabilistic modelling of the data.
As already mentioned, the approach \cite{Chow1970} investigates optimum reject options provided the true probability density function is known. 
This rule can therefore serve as a baseline provided this ground truth is available.
In the limit case, this reject strategy provides a bound for any other measure  in the sense of the error-reject trade-of, as proved in \cite{Hansen1994ErrorReject}.
Hansen et al.\ also extent Chows's rule \cite{Chow1970} to near optimal classifiers on finite data sets, and they introduce a general scaling to compare error-reject curves of several independent experiments even with different classifiers or data sets. 
The work as presented in \cite{Fumera} also directly builds on \cite{Chow1970} and more closely investigates the decomposition of data into different regions as concerns the given classes and potential errors. 
They propose a strategy which is based on class related thresholds for more flexibility and better results in practice.
The setting that reliable class probabilities are unavailable and only empirical estimations thereof are available, is addressed in the approach \cite{herbei:2005}.

Due to this theoretical background, many approaches follow the roadmap to empirically estimate the data distribution first.
Often, GMMs are used for this purpose \cite{Devarakota136confidence,ICPR00VOLII48}.
Devarakota et al. \cite{Devarakota136confidence} extend a GMM to estimate the insecurity of a particular class membership for novel, previously unseen patterns of a new class; this estimation can yield to a reliable outlier  reject option.
Vailaya \& Jain \cite{ICPR00VOLII48} investigate the suitability of GMMs for both, rejection of outliers and ambiguous data.
In particular, they propose an efficient strategy how to determine suitable reject thresholds in these cases.
The reliable estimation of GMMs is particularly problematic for high dimensional data.
Therefore, Ishidera et al. \cite{kanji} propose a suitable approximation of the probability density function for high dimensionality, which is based on a low dimensional projection of the data.

These approaches while providing baselines against which to compare, do not address our setting of prototype-based multi-class classifiers.
We will rely on two ingredients for an efficient reject option:
(I) A suitable real-valued certainty measure \cite{esann, esann_special} and (II) A suitable definition of how to set a threshold for rejection. 
In most classical reject schemes as summarised above, one global threshold value is taken, and an optimum value depends on the respective costs of misclassification versus reject.
This, however, relies on the assumption of a suitable global scaling of the underlying certainty measure, an assumption which is usually not met in a given setting.
Therefore, we will focus on possibilities how to define optimum local thresholds, which release the burden of a globally appropriate scaling of the underlying certainty measure.
In particular, we will propose efficient schemes how to optimise local thresholds which are attached to the Voronoi cells given by the prototype-based model.

First we introduce prototype-based classifiers and the most relevant training schemes used in the following.
\section{Prototype-based Classifiers}\label{algos_data}
A prototype-based classifier is characterised by a set 
$W$ of $\xi$ prototypes $(\w_j,c(\w_j)) \in \R^M\times \{1,\ldots,Z\}$, whereby every 
prototype $\w$ is equipped with a class label $c(\w)$.
Classification takes place by a winner takes all rule (WTA):
Given a data point $\x$, its label becomes the label of the closest prototype
\begin{equation}
 c(\x) = c(\w_l) \text{ with } l=\argmin{\w_j\in W}d(\w_j,\x)
\label{eq:WTA}
\end{equation}
where $d$ is a distance measure; a common choice for $d$ is the Euclidean distance.
The closest prototype $\w_l$, the winner, is called the best matching unit.
Note that prototype-based models are very similar to $k$-NN classifiers \cite{knn} which stores all training data points as \emph{prototypes} and predict a label according to the closest ($k=1$) or the $k$ closest data points.
In contrast, prototype-based training models aim at a sparser representation of data by a predefined number of prototypes.
By means of the WTA rule, a prototype-based classifier decomposes the data into \emph{Voronoi cells}\/ or \emph{receptive fields}
\begin{equation}
V_j = \lbrace\x|d(\w_j,\x)\leq d(\w_k,\x),\forall\,k\neq j\rbrace ,\ j=1,\ldots,\xi ; 
\label{eq:voronoi}
\end{equation}
and it defines a constant classification on any Voronoi cell given by the label of its representative prototype.

Prototype locations are usually learned based on given data.
Assume a training data set $X$ is given with $N$ data points $(\x_i,y_i) \in \R^M \times \{1,\ldots,Z\}$.
$Z$ states the number of different classes.
The goal is to find prototype locations such that the induced classification of the data is as accurate as possible.  
Classical training techniques  are often based on heuristics such as the Hebbian learning paradigm \cite{Kohonen1989}, yielding surprisingly good results in typical model situations, see \cite{2007dynamics}.
More recent training schemes usually rely on a suitable cost function, including generalised LVQ (GLVQ) \cite{SatoYamada1995},
its extension to an adaptive matrix: 
generalized matrix LVQ (GMLVQ) \cite{SchneiderBH09a}, its local version (LGMLVQ) \cite{SchneiderBH09a} with local adaptive metrics, and statistical counterparts referred to as robust soft LVQ (RSLVQ) \cite{SeoObermayer2003}.
We will focus on GLVQ and its matrix version as a particularly efficient and powerful scheme, as well as RSLVQ as a full probabilistic model for which an explicit certainty value is directly available.

\paragraph{GMLVQ:}
The Generalized Matrix Learning Vector Quantization \cite{SchneiderBH09a} performs a stochastic gradient decent on the cost function in \cite{SatoYamada1995} with a more general metric $d_{\Lambda}$ than the standard Euclidean one.
This cost function is a differentiable function which strongly correlates to the (discrete) classification error:
\begin{equation}
 E_{\text{GMLVQ}} = \sum\limits_{i=1}^{N} \Phi \left( \frac{d_{\Lambda}^+ - d_{\Lambda}^-}{d_{\Lambda}^+ + d_{\Lambda}^-} \right)\ .
\label{eq:Cost_fun}
\end{equation}
Here, the metric $d_{\Lambda}$ is defined as general quadratic form
\begin{equation}
 d_{\Lambda}(\w,\x)=(\x-\w)^T\Lambda(\x-\w)
 \label{eq:quad_form}
\end{equation}
with a semi positive definite matrix $\Lambda$.
The value $d_{\Lambda}^+ = d_{\Lambda}(\w_j,\x_i)$ is the distance of a data point $\x_i$ to the closest prototype $\w_j$ belonging to the same class and $d_{\Lambda}^- = d_{\Lambda}(\w_k,\x_i)$ is the distance of a data point $\x_i$ to the closest prototype $\w_k$ belonging to a different class.
$\Phi$ is a monotonically increasing function, e.\,g.\ the identity or the logistic function.
The summands in this cost function are negative if and only if the classification of the corresponding point is correct, hence the costs correlate to the overall error and optimise the so-called hypothesis margin of the classifier \cite{SchneiderBH09a}.
Note that the value  $(d_{\Lambda}^+ - d_{\Lambda}^-)/(d_{\Lambda}^+ + d_{\Lambda}^-)$ is in between $(-1,0]$ for points $\x_i$ which are in the Voronoi cell of the prototype $\w_j$ corresponding to $d_{\Lambda}^+$.
A value close to $-1$ indicates that the data point $\x_i$ is very close to the prototype and the classification is very certain, while a value close to $0$ refers to points at the class boundary or outliers.

GMLVQ training is derived from these costs (\ref{eq:Cost_fun}) by a stochastic gradient descent with respect to the  prototype locations  and the metric parameters $\Lambda$.
Thereby, either a global matrix $\Lambda$ is used, or local matrices $\Lambda_j$ are adapted which induce the distance value for the Voronoi cell of prototypes $\w_j$ only:
\begin{equation}
 d_{\Lambda_j}(\w_j,\x)=(\x-\w_j)^T\Lambda_j(\x-\w_j)\ .
 \label{eq:quad_form_lok}
\end{equation}
The algorithm which refers to these local metrics (\ref{eq:quad_form_lok}) is called local GMLVQ (LGMLVQ) \cite{SchneiderBH09a}.
\paragraph{RSLVQ:}
The objective function of Robust Soft Learning Vector Quantization \cite{SeoObermayer2003} corresponds to a statistical modelling of the setting. 
It relies on the assumption that data points are generated by a GMM. 
The probability of mixture component $j$ generating data point  $\x$ is
\begin{equation*}
p(\x|j) = \frac{1}{(2\pi \sigma_j^2)^{M/2}} \cdot \exp\left(-\frac{d(\w_j,\x )}{2\sigma_j^2}\right)
\end{equation*}
This induces the mixture model
\begin{equation*}
		p(\x|W) = \sum \limits_{1\leq j \leq \xi}\  P(j) \cdot p(\x|j)
\end{equation*}
which describes the probability of having observed the (unlabelled) data.
The priors sum to one $\sum_j P(j)=1$.
Label information is incorporated into the model by enhancing every mixture
component (i.\,e.\ every prototype) with a class label.
Then the probability of having observed the labelled data is given by
\begin{equation*}
		p(\x,y| W) = \sum \limits_{j:c(\w_j)=y} P(j) \cdot p(\x|j).
\end{equation*}
The objective function of RSLVQ is defined as the log likelihood ratio of the observed data
\begin{equation*}
	\log L := \sum_{1\leq i\leq N} \log \frac{p(\x_i, y_i|W)}{p(\x_i|W)}
\end{equation*}
which corresponds to the optimisation of the likelihood of the observed class labels assuming an underlying mixture model and independence  of the data.
Training optimises these costs by means of a gradient ascend with respect to the prototype locations.
The bandwidth $\sigma_j$ is typically set identically for all mixture components, and it is treated as a meta-parameter.
There exist schemes which also adapt the bandwidth \cite{DBLP:journals/ijon/SchneiderBH10,DBLP:conf/ijcnn/SeoO06}.

\section{Rejection Strategies}\label{reject_option}
We are interested in rejection strategies for prototype-based classifiers or similar models that rely on two main ingredients:
\begin{enumerate}
\item A certainty measure which assigns a degree of certainty $r(\x)$ to every data point $\x$ indicating the certainty of the predicted class label, 
\item and a strategy how to reject a classification based on the certainty value; suitable reject strategies have to take into account that $r(\x)$ is not necessarily scaled in an easily interpretable or uniform way.
This means, the exact value $r(\x)$ does not necessarily coincide with the  statistical confidence (which would be uniformly scaled in $[0,1]$), and the scaling of the value $r(\x)$ might even change depending on the location of the data point $\x$.
\end{enumerate}
First, we shortly review suitable certainty measures $r(\x)$ before discussing optimum reject strategies based thereon.

\subsection{Certainty Measures}\label{measures}
\begin{figure}[ht]
\centering
\includegraphics[width=1\columnwidth]{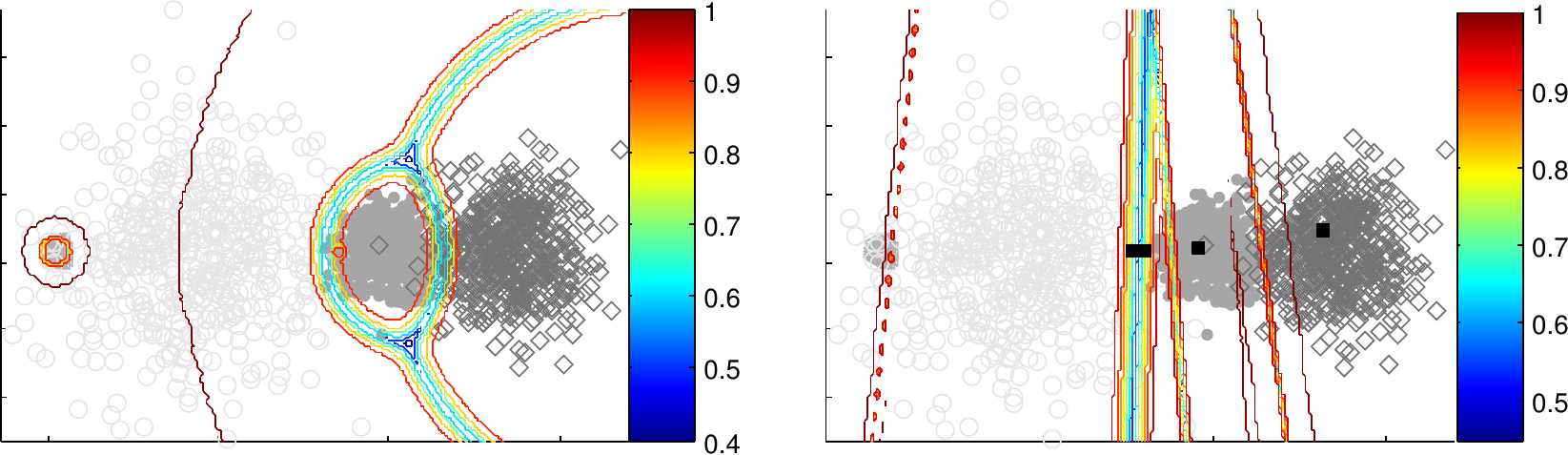} 
\caption{\cite{icann} Artificial five class data set with the contour lines of Bayes ~(\ref{eq:rejectBayes}) (left side) and the contour lines of Conf~(\ref{eq:conf}) with respect to a RSLVQ model (right side, black squares are prototypes).}
\label{fig:bayes_contour}
\end{figure}

\begin{figure}[ht]
\centering
\includegraphics[width=0.5\columnwidth]{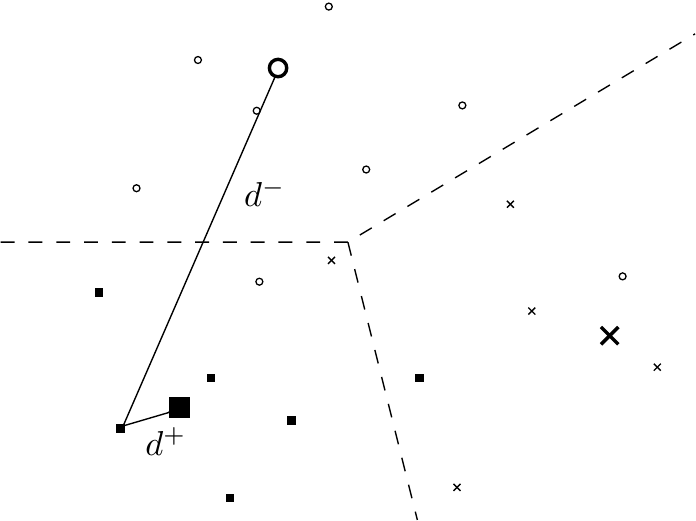} 
\caption{Sketch of an artificial three-class setting (different symbols, bigger ones are prototypes). For a single data point $d^+$, $d^-$ are shown.}
\label{fig:relsim}
\end{figure}

In the recent approaches \cite{esann, esann_special} several certainty measures have been proposed and evaluated for prototype-based classification. 
We will use three measures which scored best in the experiments as presented in \cite{esann, esann_special}: 

\paragraph*{Bayesian Confidence Value:}
Chow analysed the error-reject trade-off of Bayes classification.
He introduced an optimal certainty measure in the sense of error-reject trade-off \cite{Chow1970}. 
The certainty value for a data point $\x$ in case of a Bayes classifier is defined as:
\begin{equation}
r(\x)=\text{Bayes}(\x):=\max \limits_{1\leq j\leq Z} P(j|\x) 
\label{eq:rejectBayes}
\end{equation}
where $P(j|\x)$ is the known probability of class $j$ for a  given data point $\x$ (Fig.~\ref{fig:bayes_contour}, left).
This value can be interpreted as follows:
If the highest probability for any class with given $\x$ is lower than a defined threshold $\theta$ the probability of making a mistake is relatively high.
Classification of such data is insecure according to the chosen $\theta$.
For a binary problem the Bayes reject rule (\ref{eq:rejectBayes}) defines an interval around the decision border.
\paragraph*{Empirical Estimation of the Bayesian Probability:}
Probabilistic models like the RSLVQ model provide explicit estimations of the probability of class $j$ given a data point $\x$.
We refer to these empirical estimates as $\hat{P}(j|\x) $ and they induce a certainty measure of the form
\begin{equation}
r(\x)=\text{Conf}(\x)=:\max \limits_{1\leq j\leq Z} \hat{P}(j|\x) \ .
\label{eq:conf}
\end{equation}
An exemplary result of this measure shows Fig.~\ref{fig:bayes_contour} (right).
%
\paragraph*{Relative Similarity:}
The relative similarity (\relsim) has been proposed as a certainty measure closely related to the GMLVQ cost function~(\ref{eq:Cost_fun}), see~\cite{SatoYamada1995,esann}.
It relies on the normalised distance of a data point $\x$ to the closest prototype $d^+$ and the distance of $\x$ to a closest 
prototype of a different class $d^-$  (Fig.~\ref{fig:relsim}): 
\begin{equation}
 r(\x)= \RelSim(\x) = \frac{d^- - d^+}{d^- + d^+}
\label{eq:RelSim}
\end{equation}
whereby $d$ is the distance measure of the used algorithm ($d_{\Lambda}$ \eqref{eq:quad_form} or $d_{\Lambda_j}$ \eqref{eq:quad_form_lok}).
Note that the prototype which belongs to $d^+$ also defines the class label of $\x$.
The certainty measure \relsim~ranges in the interval $[0,1)$ where values near 1 indicate a certain classification and values near 0 are an indicator for very uncertain class labels. 

\begin{figure}[ht]
\centering
\includegraphics[width=1\columnwidth]{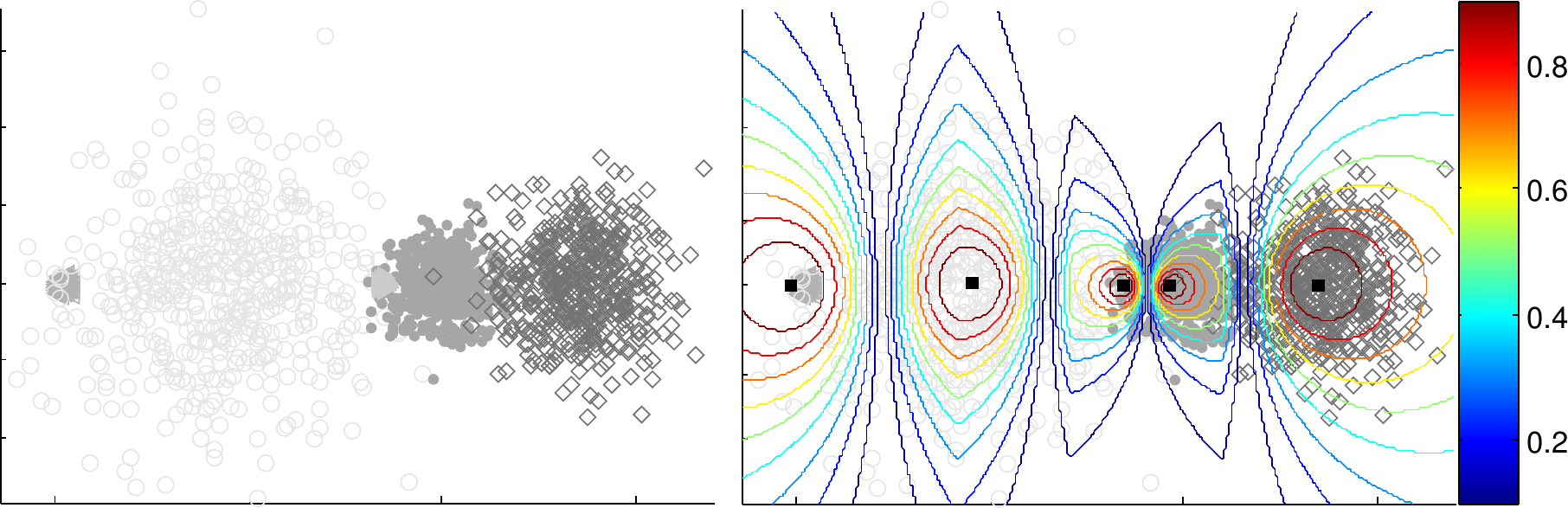} 
\caption{\cite{icann} Artificial five class data set with prototypes trained by GMLVQ (black squares) without metric adaptation. The coloured curves are the contour lines of RelSim~(\ref{eq:RelSim}).
Note that a critical region for a global threshold is between the second and the third cluster from left.
The third cluster needs a high threshold because the data points are very compact.
Applying the same threshold for the second cluster would reject most data points in this cluster which is not optimal.}
\label{fig:relsim_contour}
\end{figure}

The values $d^+$ and $d^-$ are calculated within GLVQ training schemes, hence no additional computational costs are caused by this certainty measure for training set data.
Furthermore RelSim (\ref{eq:RelSim}) depends on the stored prototypes $W$ only.
Therefore no additional storage is needed when computing the certainty of a new unlabelled data point $\x$.
Figure~\ref{fig:relsim_contour} shows the contour lines of RelSim~(\ref{eq:RelSim}) for an artificial five class problem with trained prototypes by the GMLVQ without metric adaptation, i.\,e.\ $\Lambda_{ii}=1$ and $\Lambda_{ij}=0,\ i\neq j$.
The certainty values near the class borders are low, hence the measure correctly identifies ambiguous classifications.
In addition, the contour lines have a circular shape, such that the certainty measure also correctly identifies outliers which have a large distance from the learned prototypes.

\subsection{Global Reject Option}
A global reject option extends a certainty measure by a global threshold  for the whole input space.
Assume that
\begin{equation}
	r(\x): \R^M\rightarrow\R,\ \x\mapsto r(\x)
	\label{eq:allg_measure}
\end{equation}
refers to a certainty measure where a higher value indicates higher certainty.
Given a real-valued threshold $\theta\in\R$, a data point $\x$ is rejected if and only if
\begin{equation}
r(\x)<\theta\ .
\label{eq:allg_rejection}
\end{equation}

\begin{figure}[ht]
\centering
\includegraphics[width=1\columnwidth]{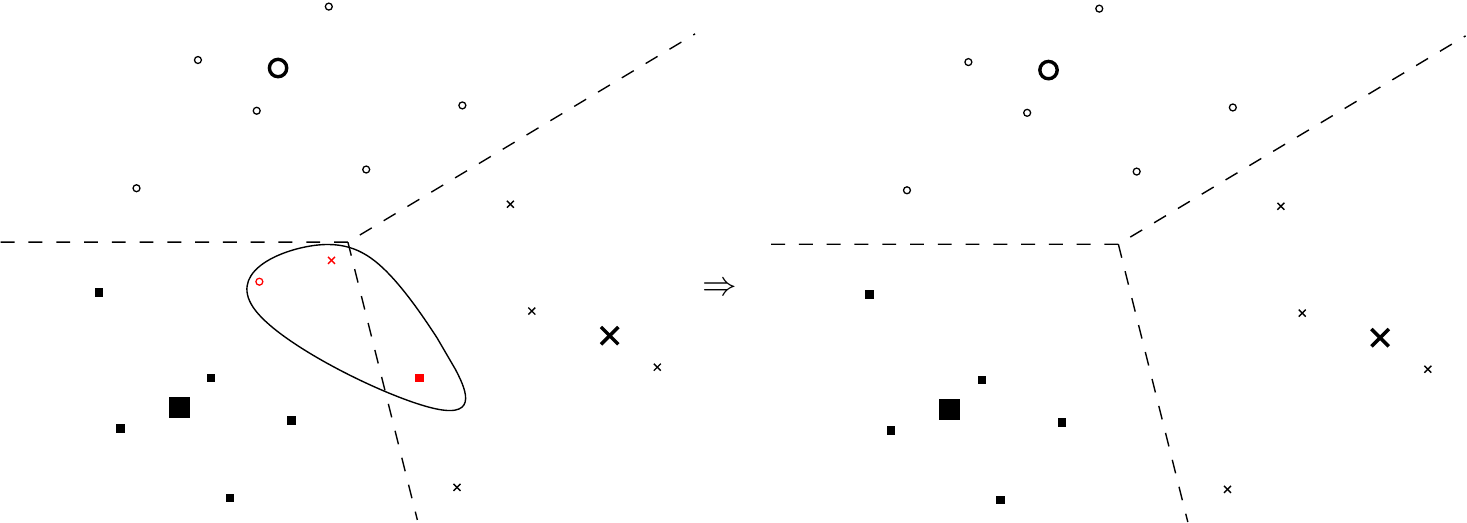}
\caption{Sketch of an artificial three-class setting (different symbols, bigger ones are prototypes) with a reject option. Left: Original model without rejection, three marked points are errors in classification. Right: Model with optimal rejection since only the three errors are rejected.}
\label{fig:opt_reject}
\end{figure}

An example of this rejection strategy is shown in Fig.~\ref{fig:opt_reject}. 
The reject option operates optimally if only labelling errors are rejected.
In general this is not the case and a reject measure leads to the rejection of a few correctly classified data points together with errors.
For optimum rejects, the number of  \emph{false} rejects should be as small as possible, while rejecting as many as possible  \emph{true} rejects. %

\subsection{Local Reject Option}

Global reject options rely on the assumption that the scaling of the certainty measure $r(\x)$ is the same for all inputs $\x$.
This assumption can be weakened by introducing local threshold strategies.
A local threshold strategy relies on a partitioning of the input space into several regions and a different choice of the reject threshold for every region; this way, it enables a finer control of rejection~\cite{ICPR00VOLII48,icann}.
Following the suggestion in \cite{ICPR00VOLII48}, we use the natural decomposition of the input space into the Voronoi-cells $V_j$ as introduced in Eq.~\eqref{eq:voronoi}.
A separate threshold $\theta_j\in\R$ is chosen for every Voronoi cell, and the reject option is given by a threshold vector $\tet = (\theta_1,\ldots,\theta_{\xi})$ of the dimension $\xi$ equal to the number of Voronoi cells $V_j$.
A data point $\x$ is rejected iff
\begin{equation*}
r(\x)<\theta_j\ \mbox{\quad where }\x\in V_j\ .
\label{eq:lokal_reject}
\end{equation*}
This means the threshold $\theta_j$ determines the behaviour for the region $V_j$ only.
In the case of one prototype per class local thresholds realise a class-wise reject option.
For the example in Fig.~\ref{fig:opt_reject} a local rejection would lead to a three-dimensional threshold vector $\tet = (\theta_1,\theta_2,\theta_3)$.
\section{Optimum Choices of Reject Thresholds}
We consider ways how to set a threshold (threshold vector) optimally for a given classifier. 
Note that rejection refers to a multi-objective: 
A threshold $\theta$ (a threshold vector $\tet$) should be chosen in such a way that the rejection of errors is maximised, while the rejection of correctly classified data points is minimised. 
To formalise this fact, and a corresponding evaluation criterion, we  explain some terms which we will use later on, first.

Assume a given data set $X\ (|X|=N)$ with labelled data for evaluation.
Applying a classification algorithm, this set decomposes into a set of correctly classified data points $L$ and a set of wrongly classified data points (errors) $E$, i.\,e.\ $X=L\cup E$.
An optimum reject would reject all points $E$, while classifying all points $L$.
Naturally, this is usually not possible using a local or global reject option.
Using a global (local) reject option by applying a threshold $\theta$ (threshold vector $\tet$), the data set $X$ decomposes into a set of rejected data points $\mathcal{X}_{\theta}$ and a set of data points remaining in the system 
$X_{\theta}$, i.\,e.\ $X=\mathcal{X}_{\theta}\cup X_{\theta}$.
We refer to data points 
\begin{equation*}
\mathcal{L}_{\theta}=\mathcal{X}_{\theta}\cap L
\end{equation*} 
as \textit{false rejects} because the rejection of correctly classified data points is undesired.
The rejection of errors 
\begin{equation*}
\mathcal{E}_{\theta}=\mathcal{X}_{\theta}\cap E
\end{equation*}
is desired therefore we call them \textit{true rejects}. 
Obviously, we can decompose $\mathcal{X}_{\theta}= \mathcal{E}_{\theta} \cup \mathcal{L}_{\theta}$.

For an evaluation, we want to report the accuracy of the obtained classifier, taking the rejected points into account.
This multi-objective can be evaluated by a reference to the so-called  accuracy reject curve (ARC) \cite{NadeemZH10}.  
For a given threshold $\theta$ (threshold vector $\tet$), this counts the accuracy of the classified points 
\begin{equation}
t_a(\theta):=(|L|-|\mathcal{L}_{\theta}|)/|X_{\theta}|
\label{arc_a}
\end{equation}
versus the ratio of the classified points
\begin{equation}
t_c(\theta):=|X_{\theta}|/|X|\ .
\label{arc_b}
\end{equation}
These two measures quantify contradictory objectives with limits
$t_a(\theta)=1$ and $t_c(\theta)=0$ for large $\theta$ (all points are rejected) and $t_a(\theta)=|L|/|X|$ and $t_c(\theta)=1$ for small $\theta$ (all points are classified, the accuracy equals the accuracy of the given classifier for the full data set).
We are interested in thresholds, such that the value $t_a$ is maximised, and $t_c$ is minimised.
Hence, not all possible thresholds and corresponding pairs $(t_a(\theta),t_c(\theta))$ are of interest, but optimum choices only, which correspond to the so-called Pareto front.
Note that pairs $(|\mathcal{L}_{\theta}|,|\mathcal{E}_{\theta}|)$ uniquely correspond to pairs $(t_a(\theta),t_c(\theta))$ and vice versa.

Every threshold uniquely induces a pair $(|\mathcal{L}_{\theta}|,|\mathcal{E}_{\theta}|)$ and a pair $(t_a(\theta),t_c(\theta))$.
We say that  $\theta'$ \emph{dominates} the choice $\theta$ if $|\mathcal{L}_{\theta'}|\le|\mathcal{L}_{\theta}|$ and $|\mathcal{E}_{\theta'}|\ge |\mathcal{E}_{\theta}|$ and for at least one term, inequality holds.
We aim at the \emph{Pareto front}
\begin{equation}
{\cal P}_{\theta}:=\{(|\mathcal{L}_{\theta}|,|\mathcal{E}_{\theta}|)|\:|\: \theta\mbox{ is not dominated by any }
\theta'\}\ .
\label{pareto}
\end{equation} 
Every dominated threshold (threshold vector) corresponds to a sub optimum choice only: 
We can increase the number of true rejects without increasing the number of false rejects, or, conversely, false rejects can be lowered without lowering true rejects.

To evaluate the efficiency of a threshold strategy, it turns out that a slightly different set is more easily accessible.
We say that  $\theta'$ \emph{dominates} $\theta$ \emph{with respect to the true rejects} if $|\mathcal{L}_{\theta'}|=|\mathcal{L}_{\theta}|$ and $|\mathcal{E}_{\theta'}|> |\mathcal{E}_{\theta}|$. 
This induces the \emph{pseudo Pareto front}
\begin{align}
\hat{\cal P}_{\theta}:=\{(|\mathcal{L}_{\theta}|,|\mathcal{E}_{\theta}|)|\:|\: 
\theta\mbox{ is not dominated by any}\\\notag
\theta' \mbox{ with respect to the true rejects}\}\ .
\label{pseudopareto}
\end{align}
Obviously, ${\cal P}_{\theta}$ can easily be computed as the subset of $\hat{\cal P}_{\theta}$ by taking the minima over the false rejects. $\hat{\cal P}_{\theta}$ has the benefit that it can be understood as a graph where  $|\mathcal{L}_{\theta}|$ varies in between $0$ and $|L|$
and $|\mathcal{E}_{\theta}|$ serves as function value.
Having computed $\hat{\cal P}_{\theta}$ and the corresponding thresholds, we report the efficiency of a rejection strategy by the corresponding ARC curve, i.\,e.\ the pairs$(t_a(\theta),t_c(\theta))$: 
These pairs correspond to a graph, where we report the ratio of classified points (starting from a ratio $1$ up to $0$) versus the obtained accuracy for the classified points. For good strategies, this graph should be increasing as fast as possible.
In the following, we discuss efficient strategies to compute the pseudo Pareto front for global and local reject strategies.

\subsection{Optimum Global Rejection}
\begin{figure}[ht]
\centering
\includegraphics[width=1\columnwidth]{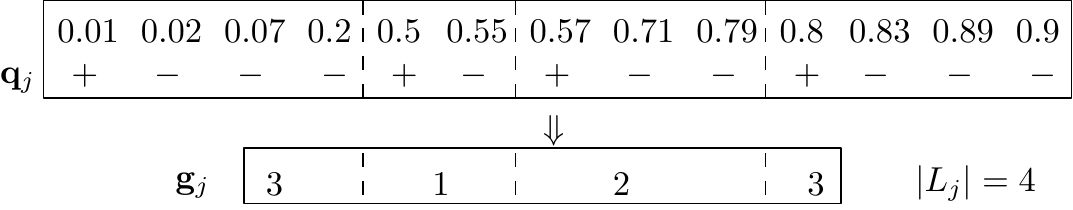} 
\caption{Reject thresholds for an area with $13$ data points. 
The first row reports the sorted certainty values $r(\x_i)$, the second row encodes if a data point is correct ($+$)/wrong ($-$) classified. 
In this case, there are $4$ thresholds which correspond to the Pareto front, according to the number of signs $+$ (taking into account the fact that point $13$ is in $E$).
The third row shows the gain $g$ when increasing the threshold value
$\theta$.
}
\label{fig:bsp}
\end{figure}

For a global reject option, only one parameter  $\theta$ is chosen.
 $|\mathcal{L}_{\theta}|$ and $|\mathcal{E}_{\theta}|$ are monotonically increasing with  increasing $\theta$, and $|X_{\theta}|$ is decreasing.
We can compute thresholds which lead to the pseudo Pareto front and the corresponding pairs $(t_a(\theta),t_c(\theta))$ in time ${\cal O}(N\log N)$ due to the following observation:
Consider the rejection measure $r(\x_i)$ as induced by the certainty function \eqref{eq:allg_measure} for all points $\x_i\in X$ and sort the values $r(\x_{i_1})<\ldots <r(\x_{i_{N}})$ (see Fig.~\ref{fig:bsp}).
Additionally Fig.~\ref{fig:bsp} indicates via the symbol 
$q_j\in\{+,-\}$ whether the corresponding point is in $L$ or in $E$.
We assume that the certainty values are not exactly identical, for simplicity;
otherwise, we sort the points such that the points in $L$ come first.
The following holds:
\begin{itemize}
\item
Every pair $(|\mathcal{L}_{\theta}|,|\mathcal{E}_{\theta}|)\in \hat{\cal P}_{\theta}$ is generated by some $\theta=r(\x_{i_j})$  which corresponds to a certainty value in this list or which corresponds to $\infty$ (i.\,e.\ all points are rejected), since values in between do not alter the number of rejected points on $X$.
\item
Values $r(\x_{i_k})$ with $\x_{i_k}\in  E$ are dominated by  $r(\x_{i_{k+1}})$ (or $\infty$ for the largest value)  with respect to true rejects since  the latter threshold accounts for the same number of false rejects, adding one true reject $\x_{i_k}$.
\item
	Contrary, values $r(\x_{i_k})$ with $\x_{i_k}\in L$ are not dominated with respect to the number of true rejects.
Increasing this threshold  always increases the number of false rejects by adding $\x_{i_k}$ to the rejected points.
\end{itemize}
Hence, the pseudo Pareto front is  induced by a set of thresholds $\Theta$  corresponding to correctly classified points:
\begin{equation}
\begin{aligned}
\Theta:=\{\theta=r(\x_{i_k})\:|\: \x_{i_k}\in L\}
\mbox{}\cup\{\infty\:|\:\text{if } x_{i_{N}}\not\in L\}\ .
\end{aligned}
\label{paretotheta}
\end{equation}
Obviously, $|\Theta|\in\{|L|,|L|+1\}$ depending on whether the last point in this list is classified correctly or not.
An exemplary setting is depicted in Fig.~\ref{fig:bsp}.
We refer to the thresholds obtained this way as $\theta(0),\ldots,\theta(|\Theta|-1)$ whereby we assume that these values are sorted in ascending order.

In addition, we can compute the gain $|g(k)|$ which is obtained when increasing the threshold from $\theta(k-1)$ to $\theta(k)$:
For $k=0,\ldots,|\Theta|-1$, the quantity
\begin{equation}
g(k):=\{\x_i \:|\: \theta(k-1)\le r(\x_i)<\theta(k), \x_i\in E\} 
\label{gain}
\end{equation}
denotes the set of additional true rejects when increasing  $\theta$ from $\theta(k-1)$ to the value $\theta(k)$ where we define $\theta(-1):=0$.
Note that $|g(0)|$ equals the maximum number of true rejects without any false reject.
It can easily be computed  by one scan through the sorted list of certainty values, see Fig.~\ref{fig:bsp}.
Obviously, the set
\begin{equation}
{\mathcal E}_{\theta(k)} = \bigcup_{0\le i\le k} g(i),\ \ k=0,\ldots,|\Theta|-1
\end{equation}
describes true rejects for the choice $\theta:=\theta(k)$.
Note that the loss due to an increase of the threshold from $\theta(k)$ to $\theta(k+1)$ is always one, by adding exactly one false reject, i.\,e.\ $|\mathcal{L}_{\theta(k)}|=k$.

\subsection{Optimum Local Rejection}
Finding the pseudo Pareto front for local rejection is more difficult than for a global one because the number of parameters (thresholds) in the optimisation rises from one to $\xi$.
First, we will derive an optimal solution via dynamic programming (DP) \cite{dp,dp2}.
Secondly, we will introduce a faster greedy solution which provides a good approximation of DP.

For every single Voronoi cell $V_j$, the optimum choice of a threshold and its corresponding pseudo Pareto front is given in exactly the same way as for the global reject option:
We sort the certainty values of the points in this Voronoi cell and look for the thresholds induced by correctly classified points (possibly adding $\infty$) as depicted in Fig.~\ref{fig:bsp}. 
We use the same notation as for a global reject option, but indicate via an additional index $j\in\{1,\ldots, \xi\}$ that these values refer to Voronoi cell $V_j$:
The correctly classified data points in $V_j$ are $L_j := L \cap V_j$, misclassified points are $E_j :=  E \cap V_j$.
A threshold $\theta_j$ in $V_j$  leads to false and true rejects $\mathcal{L}^j_{\theta_j}$ and  $\mathcal{E}^j_{\theta_j}$, respectively.
These rejects accumulate as $\mathcal{L}_{\tet} = \cup_j \mathcal{L}^j_{\theta_j}$ and $\mathcal{E}_{\tet} = \cup_j \mathcal{E}^j_{\theta_j}$ over the entire classifier,  characterising the false and true rejects of the reject strategy with threshold vector $\tet$. 
For any separate Voronoi cell, optimum thresholds as concerns the number of true rejects are induced by the certainty values of correctly classified points in this Voronoi cell, possibly adding $\infty$.
These thresholds are referred to as
\begin{equation}
\Theta_j:=
\{\theta_j(0),\ldots,\theta_j(|\Theta_j|-1)\}
\end{equation}
equivalent to  (\ref{paretotheta})
for Voronoi cell $V_j$ only,
where $|\Theta_j|\in\{|L_j|,|L_j|+1\}$.
These thresholds lead to gains $|g_j(k)|$ equivalent to (\ref{gain}) but restricted to Voronoi cell $V_j$, with true rejects  $\mathcal{E}^j_{\theta_j(k)}=\sum_{i\le k} g_j(i)$ and false rejects $\mathcal{L}^j_{\theta_j(k)}$.

We are interested in threshold vectors which describe the pseudo Pareto front of the  overall strategy,
 i.\,e.\ parameters $\tet$ such that no $\tet' \neq\tet$ exists which dominates $\tet$ with respect to the true rejects. 
 Obviously, the following relation holds:
$\tet$ is optimal $\Rightarrow$ every $\theta_j$ is optimal in $V_j$.
 Otherwise, we could easily improve $\tet$ by improving its suboptimal component.
The converse is not true:
As an example, assume Voronoi cells and thresholds as shown in Table~\ref{tab:tab1}.
Here, we can compare the threshold vectors $(1,1,1)$ and $(0,0,3)$.
While both choices lead to $3$ false rejects, the first one encounters $9$ true rejects and the second one leads to $25$ true rejects.
Hence the second vector dominates the first one with respect to true rejects, albeit all threshold components are contained in the pseudo Pareto front of the corresponding Voronoi cell.
 
 \begin{table}
\caption{Example rejects for three Voronoi cells and their losses/gains (global).}
\label{tab:tab1}
\centering
\begin{tabular}{c|cccc|cccc} 
\toprule 
threshold $i$ & 0 & 1 & 2 & 3 & 0 & 1 & 2 & 3 \\ 
 \midrule
&\multicolumn{4}{l}{$|{\mathcal L}^j_{\theta_j(i)}|$}&\multicolumn{4}{|l}{$|{\mathcal E}^j_{\theta_j(i)}|$}\\ 
\midrule 
$V_1$    & 0 & 1 & 2& 3 &   3&4&6&9\\
$V_2$  & 0 & 1 & 2&-   & 2&3&6&-\\
$V_3$ & 0 & 1 & 2 & 3&   1&2&10&20\\ 
\bottomrule
\end{tabular}
\end{table}
 
Hence we are interested in efficient strategies that compute the set of optimum threshold vectors as combinations of the single values in $\Theta_j$.
There exist at most $|\Theta_1|\cdot\ldots\cdot|\Theta_{\xi}| = {\cal O}(|L|^{\xi})$ different combinations (using the trivial upper bound ${\cal O}(|L_j|)\le {\cal O}(|L|)$ for each $|\Theta_j|$, we can expect an order ${\cal O}(|L|/\xi)$ provided the Voronoi cells have roughly the same size).
While it is possible to  test all possibilities provided a low number of prototypes $\xi$ is present, this number is infeasible if the number of prototypes gets large; this is the case in particular in online schemes or applications for big data.
In the following, we propose two alternative methods to compute the Pareto front that are linear with respect to $\xi$.

\subsubsection{Local Threshold Adaptation by DP}
For any number $0\le n\le |L|$, $1\le j\le \xi$, $0\le i \leq |\Theta_j|-1$ we define:\vspace*{-0.1cm}
\begin{equation}
\begin{aligned}
&\opt(n,j,i):=\mbox{}\\
&\max_{\tet}\{|{\mathcal E}_{\tet}|\:|\: |{\mathcal L}_{\tet}|= n,\\
&\hspace*{1cm}\theta_k\in\{\theta_j(0),\ldots,\theta_j(|\Theta_j|-1)\}\,\forall k<j,\\
&\hspace*{1cm}\theta_j\in\{\theta_j(0),\ldots,\theta_j(i)\},\\
&\hspace*{1cm}\theta_{k}=\theta_{k}(0) \, \forall k>j\}
\end{aligned}
\label{term}
\end{equation}
The term $\opt(n,j,i)$ measures the maximum number of true rejects that we can obtain with  $n$ false rejects, and a threshold vector that is restricted in the sense that the threshold  in Voronoi cell $j$ is one of the first $i$ thresholds, it is any threshold value for Voronoi cell $k<j$, and the threshold for any Voronoi cell $k>j$ is fixed to the first threshold value.
For technical reasons, it is useful to extend the index range of the Voronoi cells with $0$ that refers to the initial case that all thresholds are set to 0 which serves as an easy initialisation.
Since there are no thresholds to pick in Voronoi cell $V_0$, we define $|\Theta_0|=1$, i.\,e.\ the index i is the constant 0 in this virtual  cell $V_0$.

For $\opt(n,j,i)$, a few properties hold: 
First, obviously the pseudo Pareto front can  be recovered from the values $\opt(n,\xi,|\Theta_{\xi}|-1)$ for $n\le |L|$, since these parameters correspond to the optimum number of true rejects provided $n$ false rejects and free choice of the thresholds.
Hence an efficient computation scheme for the quantities $\opt(n,j,i)$ allows to efficiently compute the Pareto front.

Second, the decomposition of the optimality terms along the possible threshold values gives rise to the following Bellmann optimality equation:
\begin{equation}
\begin{aligned}
&\opt(n,j,i)=\\
&\left\{
\begin{array}{ll}
\mbox{if }n=0:&\hspace{-0.2cm}\sum_{k=1}^{\xi}  |{\mathcal E}^k_{\theta_k(0)} |\\[.2em]
\mbox{if }n>0, j=0:&\hspace{-0.2cm} -\infty \\[.2em]
\mbox{if }n>0, j>0, i=0:&\hspace{-0.2cm}\opt(n,j-1,|\Theta_{j-1}|-1)\\[.2em]
\mbox{if }0<n<i, j>0: &\hspace{-0.2cm}\opt(n,j,i-1)\\
\mbox{if }n\ge i>0, j>0 :&\\
&\hspace{-3cm}\max\{\opt(n,j,i-1),\\
&\hspace{-2.15cm}\opt(n-i,j-1,|\Theta_{j-1}|-1)\\
&\hspace{-0.5cm}+|{\mathcal E}^j_{\theta_j(i)}|-|{\mathcal E}^j_{\theta_j(0)}|
\}
\end{array}
\right.
\end{aligned}
\label{eq:opt_dp}
\end{equation}
This recursion captures the decomposition of the problem along the Voronoi cells as follows:
\begin{itemize}
\item
In the first case, no false rejects are allowed. Therefore, the gain is characterised by the sum of the gains $|{\mathcal E}^k_{\theta_k(0)}|$ over all Voronoi cells; these gains correspond to the minimum thresholds in all Voronoi cells which do not reject a correct point.
\item
In the second case, the number of false rejects has to equal $n$, but only  a trivial threshold with no rejects is allowed.
Hence this choice is impossible, reflected in the default value $-\infty$.
\item
In the third case, the threshold of Voronoi cell $j$ and all Voronoi cells with index larger than $j$ by definition of $\opt$ \eqref{term} are clamped to the first one.
Hence, by definition of the quantity $\opt$ \eqref{term}, this is exactly the same as the term $\opt(n,j-1,|\Theta_{j-1}|-1)$ where no restriction is posed on Voronoi cells $1$ to $j-1$, but thresholds are clamped starting from Voronoi cell $j$.
\item 
In the fourth case, the threshold number $i$ is allowed, but it would account for $i$ false rejects in the Voronoi cell $j$ with only $n<i$ false rejects allowed.
Hence we cannot pick number $i$ put a smaller one only.
\item
The fifth case considers the interesting setting where optimality is non-trivial: 
The choice of threshold number $i$ in Voronoi cell $j$ is possible, but it is unclear whether it is optimum.
There are only two possible choices: 
The first is to take a threshold with smaller index in Voronoi cell $j$, the second is to choose threshold $i$ in Voronoi cell $j$.
The first choice leads to $\opt(n,j,i-1)$ true rejects. 
The second choice has the consequence, that $i$ false rejects occur in Voronoi cell $j$, hence we are only allowed to  reject at most $n-i$ additional false rejects in Voronoi cells $1$ to $j-1$.
In turn, however, there are $|{\mathcal E}^j_{\theta_j(i)}|$ true rejects in Voronoi cell $j$ as compared to only $|{\mathcal E}^j_{\theta_j(0)}|$ if we would pick the smallest threshold in this Voronoi cell without false rejects.
Hence the optimum number of true rejects which can be achieved in this case decomposes into the optimum  $\opt(n-i,j-1,|\Theta_{j-1}|-1)$ which picks the best thresholds for Voronoi cells $1$ to $j-1$, and keeps all larger ones to the smallest possible value, and the gain $|{\mathcal E}^j_{\theta_j(i)}| -|{\mathcal E}^j_{\theta_j(0)}|$ which we obtain because picking threshold number $i$ instead of the first one in Voronoi cell $j$.
\end{itemize}
This recursive scheme can be computed by DP, since, in every recursion, the value $i$ or $j$ is decreased, and the recursion does not refer to values with larger indices.
An explicit iteration scheme can be structured in three nested loops over $n\in\{0,\ldots,|L|\}$ followed by $j\in\{1,\ldots,\xi\}$ followed by $i\in\{0,\ldots,|\Theta_j|-1\}$. 
Since every evaluation of the equation \eqref{eq:opt_dp} itself is constant time, this results in a computation scheme with effort ${\cal O}(|L|\cdot\xi\cdot\max_k|\Theta_k|)$.
Memory efficiency is  ${\cal O}(|L|\cdot \max_k|\Theta_k|)$,  since the recursion for threshold $i$ in Voronoi cell $j$ refers to the value $i-1$ only, or it directly decreases $j$.
Thus a memory matrix of dimensionality ${\cal O}(|L|\cdot \xi)$ suffices.
This DP scheme yields the optimum achievable values of true rejects; one can easily compute optimum threshold vectors thereof since they correspond to the realisation of the maxima in the recursive scheme.
Hence a standard back-tracing scheme on the matrix reveals these vectors.
See Algorithm~\ref{algo:dp} for pseudo code.
For memory efficiency we reduce the tensor $\opt(n,j,i-1)$ to a matrix $\opt(n,j)$.
The value of $\opt(n,j)$ denotes the maximum number of true rejects with $n$ false rejects and flexible thresholds in Voronoi cells $1,\ldots,j$.
In this context the vector $\tet(n,j)$ defines the optimal threshold vector for $n$ false rejects and flexible threshold in the Voronoi cells $1,\ldots,j$ whereas the Voronoi cells $j+1,\ldots,\xi$ are set to the default thresholds (no true reject).

\subsubsection{Local Threshold Adaptation by an Efficient Greedy Strategy}
Albeit enabling an optimum choice of the local threshold vectors for given data, DP as proposed above \eqref{eq:opt_dp} is infeasible for large training sets since it scales quadratically with the number of data: 
The number of thresholds $\max_j|\Theta_j|$ scales with $N$, we can expect it is of order ${\cal O}(N/\xi)$. 
An even more severe bottleneck is the time complexity for DP, which is linear in the number of data points, hence it is not suitable for big data or online schemes.
Therefore, we propose a direct greedy approximation scheme which is inspired by the full DP and which yields to an (besides pre-processing) only linear method with excellent performance at the price of possible sub optimality of the solution.

The basic idea is to start with the initial setting analogical to $\opt(0,\xi,|\Theta_{\xi}|-1)$: 
All thresholds are set to the first choice $\theta_j(0)$, hence no false rejects are present and the number of true rejects can easily be computed.
Then, a greedy threshold increase is done until the number of true rejects corresponds to the maximum possible number $|E|$. 
While increasing the values, the respective optima are stored; here, we directly compute the ARC, it would easily be possible to compute the number of true and false rejects and the corresponding thresholds, instead.

The greedy step proceeds as follows:
Starting from $n=0$, in each round, the number of false rejects $n$ is increased by one (the default case) or more than  one (in case of ties, which particularly happens if the increase of a threshold does not affect the number of true rejects but increases false rejects only).
This threshold increase is always done in the Voronoi cell with maximum immediate gain. 
More precisely:
\begin{itemize}
\item We consider local gains for each Voronoi cell: These values are the numbers of true rejects gained by increasing the threshold index by one in this Voronoi cell.
In addition, we evaluate global gains, that are obtained when accumulating all false  rejects in one Voronoi cell only, and setting the other thresholds to the first one. 
All local and global gains can be computed directly.
\item If a global gain surpasses the local gains, this setting is taken and greedy optimisation continues. 
\item If a local gain surpasses the global gain, it is checked whether this choice is unique, or whether more than one Voronoi cell would allow a threshold increase with the same quality.
In the former case, this increase is carried out, and the greedy step continues.
\item Otherwise, a tie occurs; this is in particular the case when the increase of thresholds does not increase the number of true rejects: This happens, for example, if the considered threshold corresponds to a point in a cluster of correctly labelled points; then, a threshold increase only rejects points from this cluster, but no true rejects.
In this case, we allow to increase the number of false rejects until the tie is broken. 
\end{itemize}
This procedure is described in detail in Algorithm \ref{algo:algo}.
Thereby, we do not explicitly check whether the considered threshold indices are still in a feasible range; rather, we implicitly assume that the corresponding gain is set to $-\infty$ if the threshold would be infeasible.
The algorithm does not necessarily provide the optimum threshold vectors and hence an approximation to the quasi Pareto front only, but, as we will see in experiments, it is very close to it.
Unlike the exact algorithm, it works in ${\cal O}(|L|\cdot \xi)$ time and ${\cal O}(\xi)$ memory. 
\begin{table}
\footnotesize
\caption{Iterations of the greedy algorithm \ref{algo:algo}. 
It is shown how the false rejects are split to the Voronoi cells $V_j$.}
\label{tab:tabelle}
\centering
\begin{tabular}{l|cccccccccc} 
\toprule 
 false rejects & 0 & 1 & 2 & 3& 4 & 5& 6 &7& 8 \\ 
\midrule 
$V_1:$   & 0 & p & 0 & 0& p &0& 1&2&3\\
$V_2:$  & 0& a & 0 & 0&a &2& 2&2&2\\
$V_3:$ & 0& t &2 & 3& t&3&3&3&3\\ 
\midrule
true rejects &6& &15&25& &29&30&32&35\\
\bottomrule
\end{tabular}
\end{table}
One example of the algorithmic loops is depicted in Table~\ref{tab:tabelle} for the gains as shown in 
Table~\ref{tab:tab1}.
The table shows the picked threshold indices of the consecutive iterations of the greedy search.

\section{Experiments}\label{experiments}
Having proposed efficient exact and approximate algorithms to determine optimum thresholds, we evaluate the results of the reject options for different data sets.
In all cases, we use a 10-fold repeated cross-validation with ten repeats.
We evaluate the models obtained by RSLVQ, GMLVQ, and LGMLVQ with one prototype per class. 
Thereby, we can combine the models with different certainty measures depending on their output: 
Since RSLVQ provides probability estimates, we can combine it with the certainty measure Conf.
In turn, GMLVQ and LGMLVQ lend itself to the certainty measure RelSim which is computed already while training.
We compare our results with a standard rejection measure of SVM \cite{Platt99,Wu2004} which is implemented in the LIBSVM toolbox \cite{libsvm}.

For numerical reasons, we do not display the setting $|X_{\theta}|=0$.
In Fig.~\ref{fig:local_plot} to Fig.~\ref{fig:benchmarks}, we display the ARC averaged over 100 runs per data set and rejection measure.
Note that the single curves have different ranges for $|X_{\tet}|/|X|$ corresponding to different thresholds.
To ensure a reliable display, we only report those points $|X_{\tet}|/|X|$ for which at least 80 runs deliver a value.

\subsection{Data Sets}
For evaluation, we consider the following data sets:

\textbf{Gaussian Clusters:} This data set contains two artificially generated overlapping 2D Gaussian clusters with means  $\mu_x= (-4, 4.5)$,  $\mu_y =(4,0.5)$, 
and standard deviations $\sigma_x=(5.2,7.1)$ and  $\sigma_y=(2.5,2.1)$.
These points are overlaid with uniform noise. 

\textbf{Pearl Necklace:} This data set consists of five artificially generated Gaussian clusters in two dimensions with overlap. Mean values are given by 
$\mu_{y_i}=3\ \forall i,\ \mu_{x}=(2, 44, 85, 100, 136 )$,
standard deviation per dimension is given by $\sigma_x=(1, 20, 0.5, 7, 11 ),\ \sigma_x=\sigma_y$.

\textbf{Image Segmentation:} The image segmentation data set consists of $2310$ data points which contain $19$ real-valued image descriptors.
The data represent small patches from outdoor images with $7$ different classes with equal distribution such as grass, cement, etc.~\cite{uci}.

\textbf{Tecator:} The Tecator data set~\cite{tecator} consists of $215$ spectra of meat probes.
The $100$ spectral bands range from $850$\,nm to $1050$\,nm. 
The task is to predict the fat content (high/low) of the probes, which is turned into a balanced two class classification problem. 

\textbf{Haberman:} The Haberman survival data set includes $306$ instances of two classes indicating being alive for more than $5$ years after breast cancer surgery \cite{uci}.
One instance represents three features linked to the age, the year, and the number of positive axillary nodes detected.

\textbf{Coil:} The Columbia Object Image Database Library contains gray scaled images of twenty objects \cite{coil20}.
Each object is rotated in $5^{\circ}$ steps, resulting in $72$ images per object. 
The data set contains $1440$ vectors with $16384$ dimensions that are reduced with PCA \cite{matlab} to $30$.

Since ground truth is available for the first two, artificial data sets, we can use
optimum Bayesian decision as a Gold standard for comparison in these two cases.

\subsection{Comparison of DP vs.\ Greedy Optimization}
\begin{figure*}[ht]
\centering
\includegraphics{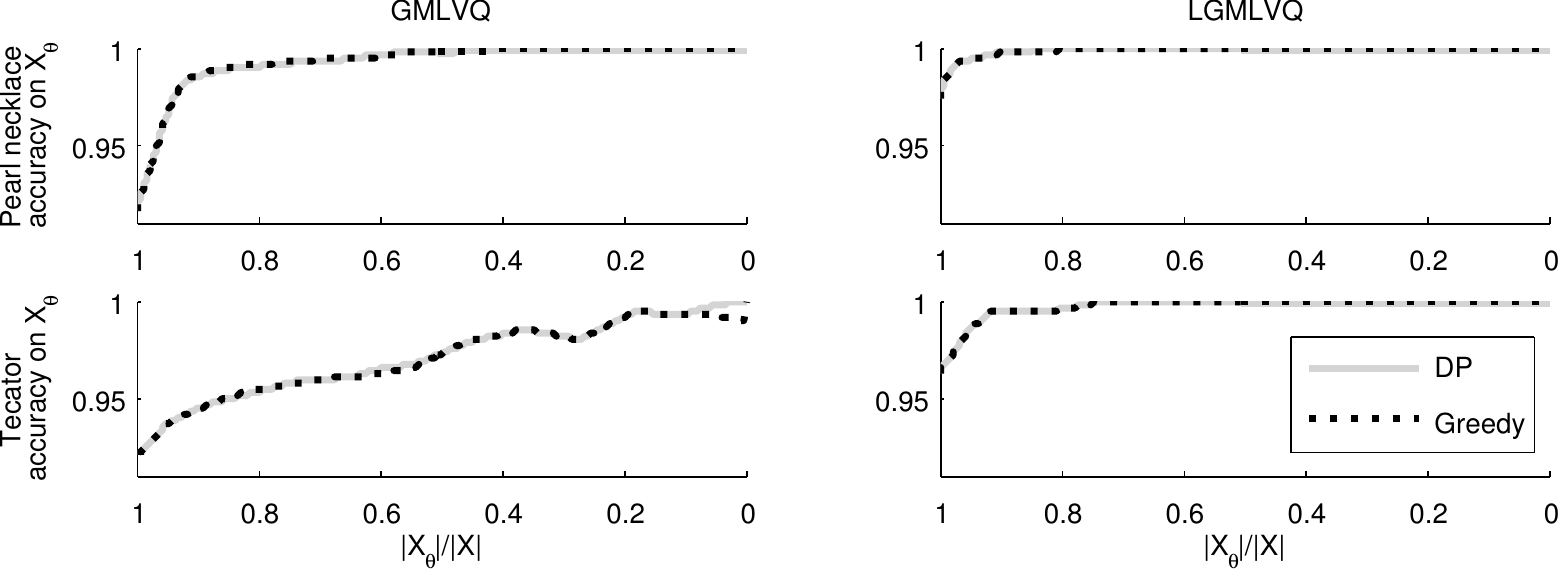} 
\caption{Averaged accuracy reject curves for dynamic programming (DP) and the greedy optimization  applied on artificial and benchmark data sets for the relative similarity (RelSim).}
\label{fig:local_plot}
\end{figure*}

First, we evaluate the performance of a greedy optimisation for  the computation of local reject thresholds versus an optimum DP scheme.
The results are compared in Fig.~\ref{fig:local_plot}.
Since we are interested in the ability of the heuristics to approximate optimum thresholds, ARCs are computed on the training set for which the threshold values are exactly optimized using DP.

One can clearly observe that the resulting curves are very similar for the shown data sets and the models provided by GMLVQ as well as LGMLVQ.
Only for the Tecator (Haberman) data set the optimum DP solution beats the greedy strategy in a small region, in particular for settings with a large portion of rejected data points (that are usually of less interest in practice since almost all points are rejected in these settings).
Results on the other data sets show a similar behaviour.

Hence we can conclude that the greedy optimisation provides near optimal results for realistic settings, while requiring
less time and memory complexity.
Because of this fact we will use the greedy optimisation for the local reject options in the following analyses.

\subsection{Experiments on Artificial Data}
\begin{figure*}[htb]
\centering
\includegraphics[width=\textwidth]{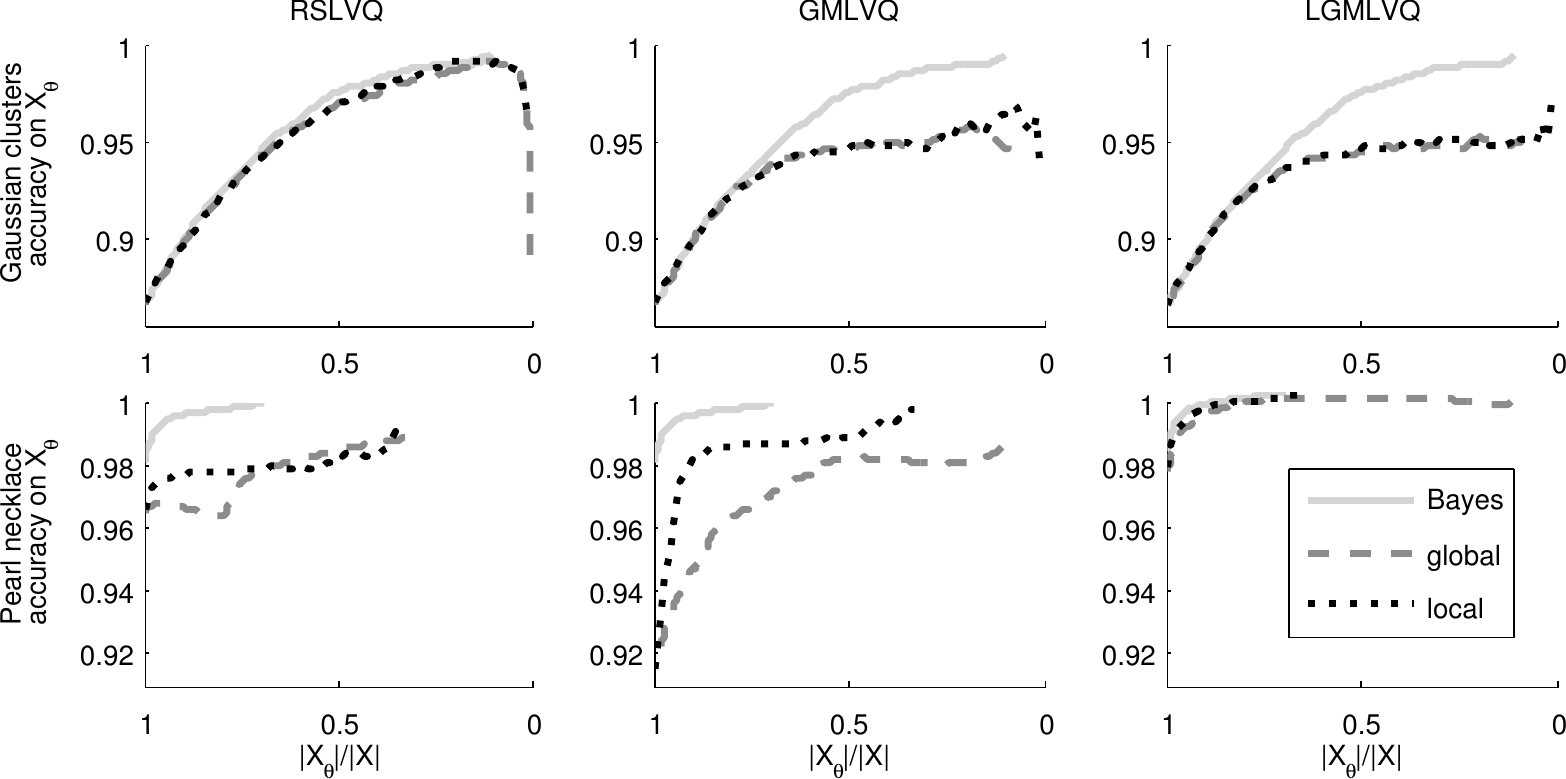} 
\caption{Averaged ARCs for global and local rejection evaluated on the test sets. For RSLVQ Conf \eqref{eq:conf} serves as rejection measure and for the other two models RelSim \eqref{eq:RelSim} serves as rejection measure. The Bayes rejection with known class probabilities provides a Gold standard for comparison.}
\label{fig:artificial}
\end{figure*}

Thereby, we report the ARC obtained on a hold out test set (which is also not used for threshold optimisation) in order to judge the interesting generalisation error of the classification models with reject option.
The data densities for the artificial data sets Gaussian clusters and Pearl necklace are known.
Hence we can compare  local and global reject options on these data with the optimum Bayes rejection, see  Fig.~\ref{fig:artificial}.
Thereby, RSLVQ is combined with Conf as rejection measure, while   RelSim is used for deterministic LVQ models, relying on the insights as gained in the studies \cite{SatoYamada1995,esann,esann_special,wsom,icann}. 
For all settings, the performance of the classifier on the test set is depicted, after optimising model parameters and threshold values on the training set.
Results of a repeated cross-validation are shown, as specified before.

\textbf{Gaussian Clusters:}
For Gaussian clusters, the global and the local rejection ARCs are almost identical for all three models.
Therefore, in this setting, it is not necessary to carry out a local strategy, but a computationally more efficient global reject option suffices.
Interestingly, reject strategies reach the quality of an optimum Bayesian reject  in the relevant regime of up to 25\,\% rejected data points as can be seen in the left part of the ARCs.
RSLVQ, due to its foundation on a probabilistic model, even enables a close to optimum rejection for the full regime, see Fig.~\ref{fig:artificial}.

\textbf{Pearl Necklace:}
The pearl necklace data set is designed to show the advantage of local rejection as already mentioned before when referring to Fig.~\ref{fig:relsim_contour}.
Here it turns out that local rejection performs better than global rejection for the models RSLVQ and GMLVQ.
As can be seen from Fig.~\ref{fig:artificial}, neither RSLVQ nor GMLVQ reach the optimum decision quality, but the ARC curves are greatly improved when using a local instead of a global threshold strategy.
This observation can be attributed to the fact that the scaling behaviour of the certainty measure is not  the same for the full data space in these settings:
RSLVQ is restricted to one global bandwidth, similarly,
GMLVQ is restricted to one global quadratic form.
This enforces a scaling of the certainty measure which does not scale uniformly with the (varying) certainty as present in the data.
In comparison, LGMLVQ is capable of reaching the optimum Bayesian reject boundary for both, local and global reject strategies, caused by the local scaling of the quadratic form in the model.
The analysis on these artificial data sets is a first indicator that shows that local reject options can be superior to global ones in particular for simple models.
On the other side, there  might be a small difference only in between local and global reject options for good models. In all cases, a sufficiently flexible LVQ model together with the proposed reject strategies reaches the quality of an optimum Bayesian reject strategy.

\subsection{Experiments on Benchmarks}
\begin{figure*}[ht]
\centering
\includegraphics[width=\textwidth]{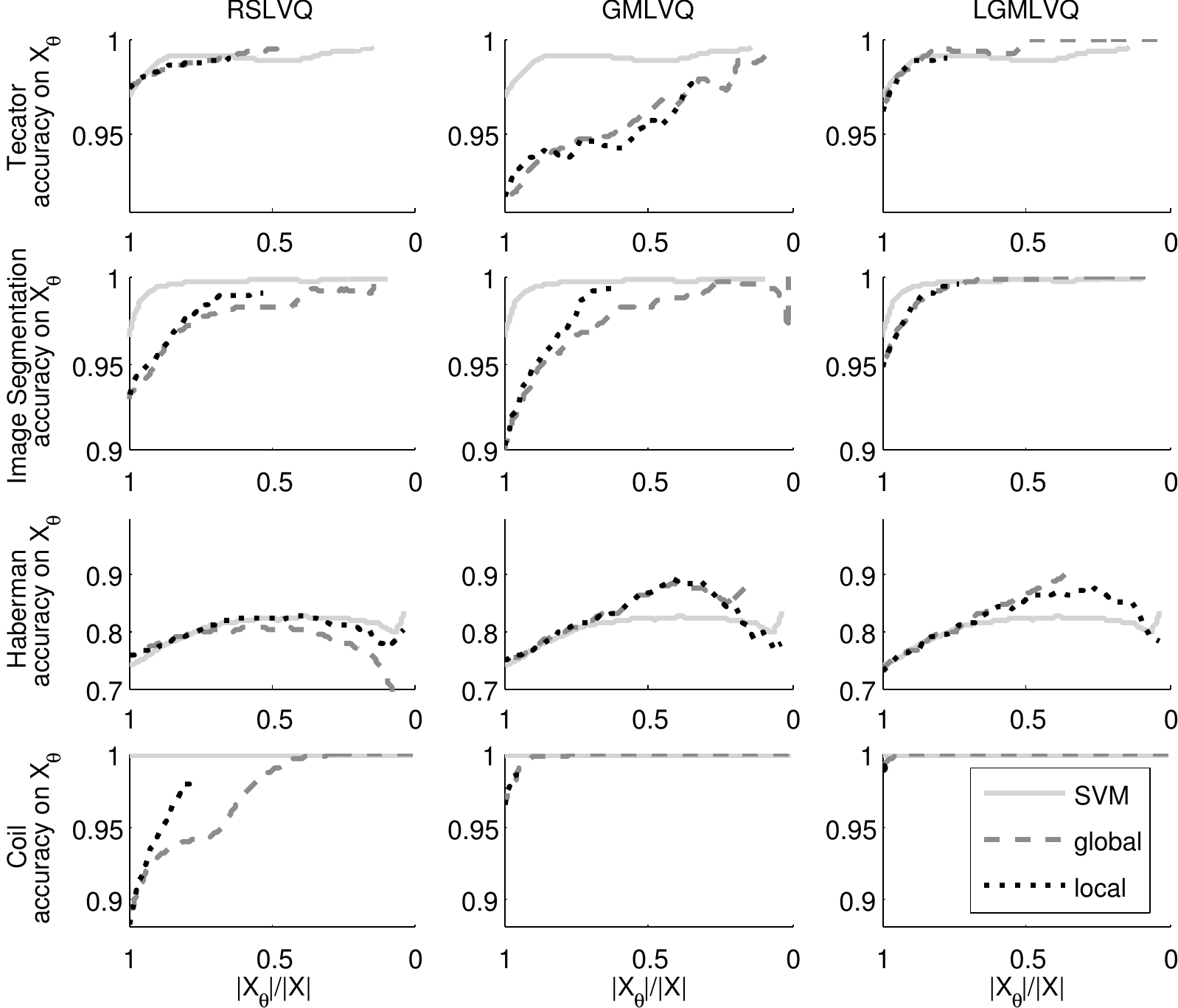} 
\caption{Averaged ARCs for global and local rejection evaluated on the test sets. For RSLVQ Conf \eqref{eq:conf} serves as rejection measure and for the other two models RelSim \eqref{eq:RelSim} serves as rejection measure. The SVM rejection is used as a state of the art method for comparison.}
\label{fig:benchmarks}
\end{figure*}

For the benchmark data sets, the underlying density models are unknown, hence we cannot report the result of an optimum Bayes rejection.
For these settings, as an alternative, we report the results which are obtained with an SVM and the reject option as introduced in \cite{Platt99,Wu2004}.
Figure~\ref{fig:benchmarks} displays all results.

\textbf{Tecator:}
RSLVQ and LGMLVQ provide results which are comparable to the SVM, while GMLVQ leads to worse accuracy. Note, however, the scaling: 
Also in the latter case, the classification accuracy of the full model is about 92\,\%, which increases to 94\,\% when rejecting 10\,\% of the data. 
For this regime for GMLVQ, the local threshold strategy is slightly better than a global one.

\textbf{Image Segmentation:}
For this setting, the SVM yields the best classification accuracy of 97\,\% compared to 95\,\% for LGMLVQ (and less for the other models).
This fact can be explained by the simpler model provided by LVQ techniques as compared to SVM, which can rely on a more complex classification boundary in this setting.
Still, the reject strategies for the LVQ models are highly performant: 
Rejecting 10\,\% of the data enables an increase of the classification accuracy by 3\,\% for LGMLVQ. 
For the simpler models RSLVQ and GMLVQ, again, a benefit of local versus global thresholds can clearly be observed.

\textbf{Haberman:}
For the Haberman data, all LVQ models display the same ARC as SVM models for the interesting regime of at most 25\% rejections in the data. For larger reject fractions, deterministic LVQ methods are superior to SVM models and corresponding reject options.

\textbf{Coil:}
The coil data set allows a high classification accuracy reaching 100\,\%. 
LVQ models display a slightly smaller accuracy for the full data set due to their simple form, representing the model by few prototypes only.
Here, the benefit of reject options is obvious, since it enables to reach 100\,\% accuracy when rejecting less than 10\,\% of the data for GMLVQ (less than 2\,\% for LGMLVQ).
The probabilistic counterpart RSLVQ performs worse, but again, the superiority of local rejects versus global options is clearly apparent for this weaker model.

Based on these experiments, we conclude the following:
\begin{itemize}
\item Reject options can greatly enhance the classification performance, provided the classification accuracy is not yet optimum.
\item Local reject options yield better results than global ones, whereby this effect is stronger for simple models for which the classification accuracy on the full data set is not yet optimum. 
For more flexible models with excellent classification accuracy for the full data set, this effect is not necessarily given.
\item LGMLVQ and the proposed reject option is comparable to SVM and the standard reject option for the considered data.
\end{itemize}
We would like to emphasise that the models as provided by LVQ techniques are sparse as compared to the SVM since we use only one prototype per class.
Further, the proposed global reject option depends on the prototypes only, while the SVM technique requires a tuning of the non-linearity on the given data \cite{Platt99}.

\subsection{Medical Application} \label{sec:medical}
We conclude  with a recent example from the medical domain.
The adrenal tumours data \cite{adrenal_esann} contains $147$ data points composed of $32$ steroid marker values.
Two classes are present: 
Patients with benign adrenocortical adenoma (ACA) or malignant carcinoma (ACC).
The $32$ steroid marker values are measured from urine samples using gas chromatography/mass spectrometry.
For further medical details we refer to \cite{adrenal_esann,arltbiehl2011jcem}.
The two classes are unbalanced with $102$ ACA and $45$ ACC data points.

Our analysis of the data follows the proposed evaluations in \cite{adrenal_esann,arltbiehl2011jcem}:
We train a GMLVQ model with one prototype per class.
We use the same pre-processing as described in \cite{adrenal_esann,arltbiehl2011jcem}.
The data set has $56$ missing values (out of 4704).  
GMLVQ can deal with these values by ignoring them for the distance computation and update, whenever the values are missing. 
This corresponds to a substitution of the values by the average as provided by the closest prototype.
The same treatment of missing values is possible when calculating the RelSim values for rejection.
For the evaluation of reject options we split the data into a train set ($90\,\%$) and a test set ($10\,\%$). 
We evaluated the ARC of $1000$ random splits of the data and the corresponding GMLVQ models.
	\begin{figure}[ht]
		\centering
		\includegraphics[width=\columnwidth]{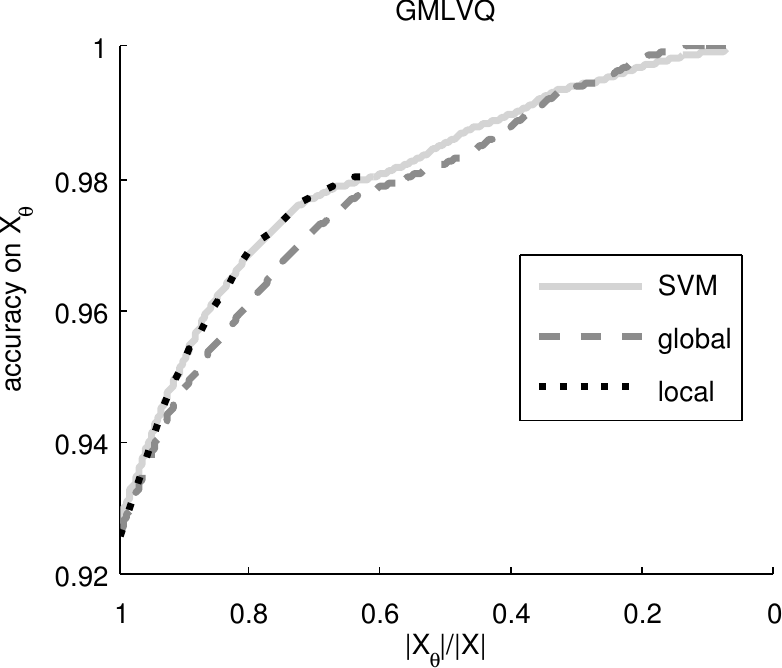} 
		\caption{Averaged ARCs for global and local rejection (test set). We use the RelSim as rejection measure.}
		\label{fig:med_apl}
	\end{figure}
The averaged ARCs of the tested reject options can be found in Fig.~\ref{fig:med_apl}.
There is nearly no difference between the curves of the global and the local rejection for small rejection rates (up to 10\,\%).
For more than 10\,\% rejection, the local rejection strategy improves the accuracy more than the global one.
Its ARC is comparable  to the ARC associated with SVM rejection computed based on LIBSVM \cite{libsvm}.
This can be attributed to the fact that the scaling of RelSim is not uniform as compared to the inherent scaling of the data in this regime.
For SVM, missing value imputation has to be done; here we replace the missing values by the class conditional means, following the suggestion in \cite{adrenal_esann}.
On average, the SVM models leads to $31$ support vectors whereas the GMLVQ models only contains $2$ prototypes. 
Further, the GMLVQ model provides insight into potentially relevant biomarkers and prototypical representatives of the classes, as has been detailed in the publication \cite{arltbiehl2011jcem}.
The suggested biomarkers, in particular, have been linked with biomedical insight \cite{arltbiehl2011jcem}.
As a conclusion, the GMLVQ model together with the proposed reject scheme offers a reliable and  compact model for this medical application.

\section{Conclusion}
In this article, we introduced reject strategies for prototype-based classifiers and extensively evaluated the proposed methods for diverse data sets, thereby comparing to state of the art reject options as present for SVM.
In particular, we introduced global and local reject strategies and addressed the problem of their efficient computation.
We introduced two algorithms to derive optimum  local reject thresholds: (i)  An optimum technique based on dynamic programming (DP) and (ii) a fast greedy approximation.
While the first is provably optimum, the latter is based on heuristics. However, we showed that the results of both solutions are very similar such that the fast greedy solution instead of the more complex solution via DP seems a reasonable choice. 
Its memory complexity is only linear with respect to the number of data, while DP requires quadratic time, and its memory complexity is constant as concerns the number of data, while DPs memory size depends linearly on the number of data points.

When investigating these techniques for diverse real-life data sets, the benefit of local strategies becomes apparent in particular for simple prototype-based models.
The effect is less pronounced for more complex models that involve local metric learning like LGMLVQ.
Interestingly, the proposed reject strategies in combination with the very intuitive deterministic method LGMLVQ lead to results which are comparable to SVM and corresponding reject options. 
Thereby, the LVQ techniques base the reject on their distance to few prototypes only, hence they open the way towards efficient techniques for online scenarios.

So far, the reject strategies have been designed and evaluated for offline training scenarios only, disregarding the possibility of trends present in life long learning scenarios, or its coupling to possibly varying costs for rejects versus errors.
We will analyse in future work how to extend the proposed methods to online scenarios and life long learning, where according thresholds are picked automatically based on the proposed results in this article.

\appendix
\begin{algorithm}[h!]
\begin{pseudocode}{DP}{\theta_j(i), {\cal E}^j_{\theta_j(i)}}
\small\vspace*{-0.1cm}
\text{// compute optimum number of true rejects by DP}\\
\small\vspace*{-0.1cm}
\text{// init}\\
h :=\sum_{k=1}^{\xi}{|\cal E}^k_{\theta_k(0)}|;\\
\FOR k:=0, \ldots, \xi \DO
		\opt(0,k):=h;\\
\FOR n:=1,\ldots, |L|\DO
\BEGIN
		\FOR k:=0, \ldots, \xi \DO
		\opt(n,k):=-\infty;
\END\\
\small
\text{// loop over number of false rejects}\\\vspace*{-0.1cm}
\FOR n:=1,\ldots, |L| \DO\hspace{-0.2cm}
\BEGIN
\small
	\text{// loop over Voronoi cells}\\
	\FOR j:=1,\ldots, \xi \DO\hspace{-0.2cm}
	\BEGIN
		 \opt(n,j):= \opt(n,j-1);\\
		 \small
		\text{//loop over thresholds in Voronoi cell j}\\
		\small
		\text{//that agree with false rejects}\\
		\FOR i:= 1, \ldots, \min\{n,|\Theta_j|-1\} \DO\hspace{-0.2cm}
			\BEGIN
			   n' := n-i;\\
			   gain := {|\cal E}^j_{\theta_j(i)}|-|{\cal E}^j_{\theta_j(0)}|;\\
				h:=\opt(n',j-1)+gain;\\
				\IF h>\opt(n,j)
				\THEN
					\opt(n,j):=h;
			\END
	\END
\END\\
\small\vspace*{-0.1cm}
\text{// compute threshold vector by back-tracing}\\
\small\vspace*{-0.1cm}
\text{// init with default value: first thresholds}\\
\FOR n:=0,\ldots,|L| \DO
\BEGIN
	\FOR k:=1\ldots \xi \DO
		\tet(n,k):=\theta_k(0);
		\END\\
		\small\vspace*{-0.1cm}
\text{// back-tracing in the matrix $\opt$}\\
\FOR n:=1,\ldots,|L| \DO
\BEGIN
	j:=\xi; \text{ // start in last Voronoi cell}\\
	n':=n;\\
	i:=\min(n', |\Theta_j|-1);\\
	\WHILE j>0 \DO
	\BEGIN
		\IF i=0 \THEN 
			\BEGIN
			\small\vspace*{-0.1cm}
				\text{// threshold $0$}\\
				j:=j-1;\\ 
				i:= \min(n', |\Theta_j|-1);
			\END
		\ELSE
			\BEGIN
				n'' := n'-i;\\
				gain := |{\cal E}^j_{\theta_j(i)}|-|{\cal E}^j_{\theta_j(0)}|;\\
				h:=\opt(n'',j-1)+gain;\\
				\IF \opt(n',j)=h
			   \THEN 
			   \BEGIN
			   \small\vspace*{-0.1cm}
			   	\text{// threshold  $i$}\\
			   	\tet(n,j):=\theta_j(i);\\
					 n':=n'';\\
					 j:=j-1;\\
					 i:=\min(n', |\Theta_j|-1);
			   \END
			  	\ELSE
				\BEGIN
				\small\vspace*{-0.1cm}
					\text{// threshold smaller}\\
					i:=i-1;
				\END
			\END
		\END
	\END\\[.2cm]
	\small\vspace*{-0.1cm}
\text{// return optimum true reject numbers}\\
\small\vspace*{-0.1cm}
\text{// and corresponding threshold vectors}\\
\RETURN{ \text{matrices } \opt(n,k) \text{ and }\tet(n,k)}
\label{algo:dp}
\end{pseudocode}
\end{algorithm}

\begin{algorithm}
\begin{pseudocode}{Greedy optimizaton}{\theta_j(i),{\cal E}^j_{\theta_j(i)}}
\label{algo:algo}
\text{// init by first thresholds} \\
\FOR j:=1,\ldots,\xi \DO \ind(j) := 0;\\
h:=\sum_{k=1}^{\xi} |{\cal E}^k_{\theta_k(0)}|\\
|{\mathcal E}_{\tet}|:= h;\\
n:=0; s:=1;\\
t_c(s) := 1-|\mathcal{E}_{\tet}|/|X|;\\
t_a(s) := |L|/(|X|- |\mathcal{E}_{\tet}|);\\
\text{// loop while true rejects can be increased}\\
\WHILE |\mathcal{E}_{\tet}|\neq |E| \DO\hspace{-0.22cm}
	\BEGIN
		\text{//most improvement locally}\\
		gain := \max_{j}\lbrace| {\cal E}^j_{\theta_j(\ind(j)+1)}| -| {\cal E}^j_{\theta_j(\ind(j))}|\rbrace;\\
		I_{gain} := \argmax{j}\lbrace |{\cal E}^j_{\theta_j(\ind(j)+1)}| - |{\cal E}^j_{\theta_j(\ind(j))}|\rbrace;\\
		\text{//most improvement globally}\\
		GAIN := \max_{j}\lbrace |{\cal E}^j_{\theta_j(n+1)}| -| {\cal E}^j_{\theta_j(0)}|\rbrace;\\
		I_{GAIN} := \argmax{j}\lbrace |{\cal E}^j_{\theta_j(n+1)}| -| {\cal E}^j_{\theta_j(0)}|\rbrace; \\
		\IF GAIN >( gain +|\mathcal{E}_{\tet}|-h)
		\THEN\hspace{-0.22cm}
			\BEGIN
				\FOR j:=1, \ldots, \xi 
				\DO \ind(j) := 0;\\
			    \ind(I_{GAIN}) := n;\\
			    |\mathcal{E}_{\tet}| := GAIN + h;\\
				 n :=n+1;\\
			\END
		\ELSE\hspace{-0.22cm}
			\BEGIN
				\IF{ I_{gain} \text{ is unique}}
				\THEN\hspace{-0.25cm}
					\BEGIN
						\ind(I_{gain}):= \ind(I_{gain}) +1;\\
						|\mathcal{E}_{\tet}| := |\mathcal{E}_{\tet}|+gain;\\
						n:= n+1;
					\END
				\ELSE\hspace{-0.22cm}
					\BEGIN 
					    \text{// increase false rejects}\\
						o :=1; \\
						\REPEAT
						\hspace{-0.25cm}\BEGIN
							o := o+1;\\
							gain :=\\
							\hspace*{.15cm} \max_{j}\lbrace| {\cal E}^j_{\theta_j(\ind(j)+o)}| -| {\cal E}^j_{\theta_j(\ind(j))}|\rbrace;\\
		I_{gain} :=\\
		\hspace*{.15cm} \argmax{j}\lbrace | {\cal E}^j_{\theta_j(\ind(j)+o)}| - |{\cal E}^j_{\theta_j(\ind(j))}|\rbrace;\\
							\END
							\UNTIL{I_{gain} \text{ is unique}};\\
						n := n +o;\\
						\ind(I_{gain}) := \ind(I_{gain})+o;\\
						|\mathcal{E}_{\tet}| := |\mathcal{E}_{\tet}|+gain;\\			
					\END
			\END\\
			s :=s+1;\\
			\mathbf{t}_c(s):=1-(n+|\mathcal{E}_{\tet}|)/|X|; \\
			\mathbf{t}_a(s):=(|L| -n)/(|X|-(n+|\mathcal{E}_{\tet}|));
	\END\\
	\RETURN{ \mathbf{t}_c,\mathbf{t}_a }
\end{pseudocode}
\end{algorithm}

\section*{Acknowledgment}
The authors would like to thank Stephan Hasler for the initial idea of the greedy optimisation of local thresholds and very helpful discussions thereon.
The authors would like to thank Wiebke Arlt and Michael Biehl for providing the adrenal tumour data and their support in related questions.

\footnotesize
\bibliographystyle{abbrv}

\bibliography{Literatur}

\end{document}